%% file: main.tex
\documentclass[sigconf,nonacm]{acmart}
\setkeys{acmart.cls}{balance=false}
\usepackage{multirow} 
\usepackage{booktabs} 
\usepackage{graphicx} 
\usepackage{tabularx}
\usepackage{makecell}
\usepackage{pifont}
\usepackage{marvosym}
\newcommand{\cmark}{\ding{51}}
\newcommand{\xmark}{\ding{55}}

\newcolumntype{C}{>{\centering\arraybackslash}X}
\usepackage{tabularray} 
\usepackage[most]{tcolorbox}
\usepackage{fontawesome5}
\UseTblrLibrary{booktabs} 
\AtBeginDocument{%
  }

\setcopyright{acmlicensed}
\renewcommand\footnotetextcopyrightpermission[1]{}
\copyrightyear{2018}
\acmYear{2018}
\acmDOI{XXXXXXX.XXXXXXX}
\acmConference[Conference acronym 'XX]{Make sure to enter the correct
  conference title from your rights confirmation email}{June 03--05,
  2018}{Woodstock, NY}
\acmISBN{978-1-4503-XXXX-X/2018/06}

\begin{document}

\title{GABench: A Comprehensive Benchmark for Evaluating LLM Agents on Graph Analysis Tasks}

% GraphAgentBench: A Comprehensive Benchmark for LLM Agents on Graph Analysis
% GraphFlowBench: A Comprehensive Benchmark for LLM Agents on Graph Analysis Workflows
% GraphFlowBench: A Comprehensive Benchmark for Evaluating LLM Agents on Graph Analysis Workflows
% A Comprehensive Graph Analysis Benchmark for Large Language Model Agents
% GraphQBA: A Comprehensive Benchmark for Evaluating LLM Agents on Graph Analysis Workflows


%% ==================== 作者与单位 ====================
% 作者列表（共同一作用 \ast，通讯作者用 \dagger 标注）
\author{Jiarui Tan$^{1\ast}$, Zhongjian Zhang$^{1\ast}$, YaBo Guo$^1$, Jiawei Liu$^2$}
\author{Yujie Xing$^1$, Muhan Zhang$^3$, Cheng Yang$^1$ and Chuan Shi$^{1\dagger}$}
\thanks{$\ast$ Both authors contributed equally to this research.\\$\dagger$ Corresponding author.}
\affiliation{%
  \institution{$^1$Beijing University of Posts and Telecommunications, $^2$The Chinese University of Hong Kong, $^3$Peking University}
  \city{}
  \country{}
}

% 邮箱区域
\email{tanjiarui@bupt.edu.cn, zhangzj@bupt.edu.cn,
gyb-15175052906@bupt.edu.cn, liujw@cse.cuhk.edu.hk}
\email{yujie-xing@bupt.edu.cn, muhan@pku.edu.cn,
yangcheng@bupt.edu.cn, shichuan@bupt.edu.cn}
\renewcommand{\shortauthors}{Jiarui Tan, Zhongjian Zhang, and YaBo Guo et al.}

\begin{abstract}
% LLM-based agent
Large language model (LLM) agents are increasingly capable of planning, using tools, and interacting with external environments. They are typically supported by harnesses, which manage state and coordinate multi-step execution.
%To support these capabilities in practice, agent harnesses provide the execution framework for tool invocation, state management, and adaptation to intermediate results.
%To evaluate these capabilities of LLM agents,  graph analysis provides a promising setting as it often requires agents to access graph data and execute graph operations in graph environment. 
Graph analysis provides a promising setting for evaluating their agentic capabilities, because it requires agents to access data and execute operations in a graph environment.
However, existing graph benchmarks for LLMs provide limited coverage of graph tasks and graph types, making it difficult to comprehensively evaluate LLM agents. 
Moreover, they typically formulate graph analysis as text-based question answering, where graph information is directly provided in the prompt, limiting the evaluation of end-to-end agentic capabilities.
%However, existing graph analysis benchmarks for LLMs provide limited coverage of graph tasks and graph types. Moreover, they typically formulate graph analysis as text-based question answering, where graph information is directly provided in the prompt.
% 
To address these limitations, we introduce \textbf{GABench}, a comprehensive benchmark for agentic graph analysis. GABench spans three graph types and covers four graph analysis task categories: graph retrieval, graph theory, graph machine learning, and graph open-ended question answering. GABench also provides 84 executable tools for accessing graph data and performing diverse graph operations. Building on these tools, we develop an agentic graph analysis task generation pipeline and construct 10,400 tasks with verifiable ground truth.
Using GABench, we evaluate a range of frontier LLMs and agent harnesses. Our experiments reveal three key findings: 
(1) Existing LLM agents still struggle with complex graph analysis tasks. (2) Harness choice significantly affects performance, yet existing harnesses remain limited on complex graph tasks. (3) Graph analysis depends more on tool-call quality than quantity.
% (1) Current LLM agents still struggle with challenging graph analysis tasks; (2) Agent harnesses substantially affect the performance of the same backbone LLM; (3) More tool calls do not necessarily lead to higher task success. 
Our findings provide practical insights into the development and evaluation of LLM agents for graph analysis. 
The data and code can be found at https://github.com/tanjjjjrr/GABench. 

\end{abstract}

\received{20 February 2007}
\received[revised]{12 March 2009}
\received[accepted]{5 June 2009}

\maketitle

\section{Introduction}

\input{introduction}
\section{Preliminaries and Problem Formulation}
\input{preliminary}
\section{Benchmark}
\input{benchmark}

\section{Experiments}
\input{experiment}

\section{Conclusion}
\input{conclusion}

\bibliographystyle{ACM-Reference-Format}
\bibliography{reference}
\clearpage

\appendix
% \section{Appendix}
\section{Related Work}
\input{related_work}

\input{Appendix}

\end{document}

%% file: Introduction.tex
% Large Language Model (LLM) agents have demonstrated remarkable capabilities in various tasks, with recent advances enabling them to interact with environments, execute tasks, and make decisions autonomously. They integrate LLMs with external tools and delicate workflows to improve reasoning and planning abilities. Accordingly, evaluating the planning and tool-use capabilities of LLM agents has become a timely research focus.~\cite{}
%Large language and vision-language models increasingly power agents that move beyond question answering to executing multi-step actions on a user’s behalf. Through Command-Line Interface (CLI)-based agent harnesses such as OpenClaw [30] and Claude Code [6], these agents plan, invoke external tools, maintain memory and state, and adapt to intermediate results across coding assistance, scientific research workflows, and everyday computer use tasks
%Large language models have increasingly evolved from simple text generators into agents that can reason, plan, use tools, and interact with external environments~\cite{openai2025gpt5,comanici2025gemini,yang2025qwen3,anthropic2025claude4}. Unlike standard question answering, where a model generates an answer directly from a given context, an agentic task requires an LLM to translate a user request into a action sequence, select appropriate tools, process intermediate observations, and adjust subsequent decisions~\cite{Yao-ICLR2025,gao2025mcpradarmultidimensionalbenchmarkevaluating,wang2025mcpbenchbenchmarkingtoolusingllm,ding2026wildclawbenchbenchmarkrealworldlonghorizon}.
Large language models (LLMs) have evolved from simple text generators into agents capable of reasoning, planning, using tools, and interacting with external environments~\cite{openai2025gpt5,comanici2025gemini,yang2025qwen3,anthropic2025claude4}. These capabilities are typically supported by agent harnesses, which provide interfaces for tool invocation, state and memory management, and iterative execution based on intermediate observations~\cite{openclaw,claudecode}. Unlike standard question answering, where a model generates an answer directly from a given context, an agentic task requires an LLM to translate a user request into an action sequence, select appropriate tools, process intermediate observations, and adjust subsequent decisions~\cite{Yao-ICLR2025,gao2025mcpradarmultidimensionalbenchmarkevaluating,wang2025mcpbenchbenchmarkingtoolusingllm,ding2026wildclawbenchbenchmarkrealworldlonghorizon}. 
As a representative agentic setting, data analysis tasks often require an LLM to access and preprocess raw data, execute data operations by writing and running analysis code, and produce a final answer~\cite{sun2026lambda,hong2025data,perez2025llm}. Accordingly, recent studies have developed data analysis benchmarks for evaluating LLM agents, attracting considerable attention and making this a timely research focus~\cite{zhang2024benchmarking,ma2026can,hu2024infiagent,gu2024blade}.

% Among these agentic tasks, data analysis is a representative setting. Completing a data analysis task often requires an LLM to access and process data, select suitable analysis methods, perform multiple intermediate operations, and produce a final answer. Accordingly, recent studies have developed data analysis benchmarks for evaluating LLM agents~\cite{sun2026lambda,hong2025data,perez2025llm,zhang2024benchmarking}. These benchmarks assess agents' capabilities in planning, tool use, and multi-step task execution.

Graphs are widely used to model complex relationships among entities in real-world systems, such as social networks, e-commerce networks, and citation networks~\cite{easley2010networks,chen2024link,xu2025graphomni}.
Analyzing these graphs is essential for understanding complex systems, uncovering latent patterns, and supporting prediction and decision-making~\cite{wei2026graphchain,bonald2020scikit,perret2019higra,peng2025graph,gross2018graph}.
To further investigate the potential of LLMs for graph analysis, recent studies have developed a series of benchmarks that evaluate their capabilities across different graph tasks and graph types. 
As summarized in Table~\ref{tab:benchmark_comparison}, most existing benchmarks focus on graph retrieval and graph theory~\cite{luo2024graphinstruct,zhang2024llm4dyg,DBLP:journals/corr/abs-2602-06319,chen2024graphwiz}, while only a few cover graph machine learning~\cite{zhang2024llm4dyg,guo2023gpt4graph,yuan2024gracore}. Meanwhile, they primarily consider numerical-attribute and text-attributed graphs~\cite{guo2023gpt4graph,DBLP:journals/corr/abs-2602-06319,yuan2024gracore}, with text-paired graphs receiving limited attention~\cite{tang2024grapharena}.

In a nutshell, existing benchmarks remain limited in their coverage of both graph tasks and graph types, making it difficult to comprehensively evaluate the graph analysis capabilities of LLM agents.
More importantly, existing benchmarks are not designed to evaluate agents' capabilities in graph analysis. They typically formulate graph analysis as text-based question answering, where graph information is provided in the prompt and LLMs directly produce a final answer without interacting with graph tools. As a result, these benchmarks do not provide an executable graph environment in which agents need to plan multiple steps, compose graph tools, and use intermediate results to complete a task.
The above limitations raise a fundamental question: \textit{Can LLM agents autonomously solve diverse graph analysis tasks through planning, tool composition, and interaction with executable graph environments?} Answering this question not only helps reveal the agentic capabilities of LLMs in graph environments, but also provides a new perspective on complex graph analysis tasks.

\input{benchmark_comparison}

To this end, we propose \textbf{GABench}, a comprehensive benchmark for evaluating LLM agents on graph analysis tasks. GABench is built on 13 real-world graph datasets spanning 6 domains and 3 graph types, and covers 4 major task categories: graph retrieval, graph theory, graph machine learning, and graph open-ended question answering. 
To support the construction of agentic graph analysis tasks at scale, GABench provides 84 executable tools for accessing graph data and performing diverse graph operations. Building on these tools, we develop an agentic task generation pipeline that grounds user requests in real-world graph data while maintaining task complexity and producing verifiable ground truth.
Based on this pipeline, GABench contains 10,400 tasks, each formulated as a practical user request and designed to be completed by accessing graph data and using executable tools.
% We use GABench to systematically evaluate representative LLM agents, leading to the following key insights:
% In summary, we make the following three contributions:

With GABench, we evaluate a range of frontier LLMs and agent harnesses to benchmark their graph analysis capabilities.
% Through extensive experiments, 
Our key insights include: (1) Existing LLM agents still struggle with complex graph analysis tasks, particularly graph machine learning and graph open-ended question answering, with success rates below 40\% in most cases, revealing their limited ability to understand structures and reason over graph data. 
(2) Harnesses significantly affect the graph analysis performance of LLM agents, but existing harnesses remain limited on complex graph tasks.
(3) Tool-call quality is more important than tool-call quantity for successful graph analysis. Stronger LLMs can typically achieve better performance with fewer tool calls than others.
% (1) Current LLM agents still struggle with challenging graph analysis tasks across all evaluated harnesses, with success rates below 30\% for graph open-ended question-answering tasks. (2) The same backbone LLM exhibits substantially different performance across agent harnesses, with success rates differing by more than 60\% on graph machine learning tasks. (3) More tool calls do not necessarily lead to higher success rates; stronger models can often achieve better performance with fewer tool calls than other models.
%First, graph open-ended question answering is the most challenging task category, with both lower task selection accuracy and success rate than graph retrieval, graph theory, and graph machine learning. Second, the choice of agent harness has substantial impact on performance, Claude Code consistently outperforms Hermes and OpenClaw on graph machine learning and graph open-ended question answering. Third, longer tool call chains do not necessarily lead to higher success rate, LLM agents can achieve better performance with fewer but more effective tool calls.
Our contributions are summarized as follows:

% In summary, we make the following three contributions:
% \begin{itemize}
\noindent $\bullet$ We introduce GABench, the first comprehensive benchmark for agentic graph analysis, covering three graph types and four task categories, and enabling systematic evaluation of LLM agents on graph analysis tasks under unified experimental settings.

\noindent $\bullet$ We propose a graph analysis question generation pipeline that combines template-based and LLM-based generation methods, producing 10,400 agentic questions grounded in real-world graph data.

\noindent $\bullet$ We conduct extensive experiments across four graph analysis task categories and three graph types, evaluating various LLMs under OpenClaw and further performing cross-harness comparisons.

%% file: benchmark_comparison.tex
\begin{table*}[t]
    \centering
    \vskip -0.05in
    \caption{Comparison of GABench with representative graph-oriented benchmarks in task coverage, supported graph types, agentic interaction, and reported graph-size range. Check marks and crosses indicate whether each capability is supported.}
    \vskip -0.15in
    \label{tab:benchmark_comparison}

    \small
    \renewcommand{\cmark}{\textcolor{green!60!black}{\ding{51}}}
    \renewcommand{\xmark}{\textcolor{red!75!black}{\ding{55}}}
    \setlength{\tabcolsep}{1.2pt}
    \renewcommand{\arraystretch}{1.0}
    \renewcommand{\tabularxcolumn}[1]{m{#1}}

    \begin{tabularx}{\textwidth}{
        >{\raggedright\arraybackslash}p{2.7cm}
        >{\centering\arraybackslash}p{1.1cm}
        >{\centering\arraybackslash}p{0.85cm}
        >{\centering\arraybackslash}p{1.8cm}
        >{\centering\arraybackslash}p{2.3cm}
        >{\centering\arraybackslash}p{1.95cm}
        >{\centering\arraybackslash}p{1.75cm}
        >{\centering\arraybackslash}p{1.4cm}
        >{\centering\arraybackslash}p{0.8cm}
        >{\centering\arraybackslash}X
    }
        \toprule

        \multirow[c]{2}{*}[-6pt]{\textbf{Benchmark}}
        & \multicolumn{4}{c}{\raisebox{1pt}{\textbf{Graph Analysis Task}}}
        & \multicolumn{3}{c}{\raisebox{1pt}{\textbf{Graph Type}}}
        & \multirow[c]{2}{*}[-6pt]{\textbf{Agentic}}
        & \multirow[c]{2}{*}[-6pt]{\textbf{Scalability}}
        \\[-2pt]

        \cmidrule(lr){2-5}
        \cmidrule(lr){6-8}

        & \makecell{\textbf{\footnotesize Graph}\\\textbf{\footnotesize Retrieval}}
        & \makecell{\textbf{\footnotesize Graph}\\\textbf{\footnotesize Theory}}
        & \makecell{\textbf{\footnotesize Graph Machine}\\\textbf{\footnotesize Learning}}
        & \makecell{\textbf{\footnotesize Graph Open-ended}\\\textbf{\footnotesize Question Answering}}
        & \makecell{\textbf{\footnotesize Numerical-}\\\textbf{\footnotesize Attribute Graphs}}
        & \makecell{\textbf{\footnotesize Text-Attributed}\\\textbf{\footnotesize Graphs}}
        & \makecell{\textbf{\footnotesize Text-Paired}\\\textbf{\footnotesize Graphs}}
        &
        &
        \\[-1pt]

        \midrule

        GraphOmni \cite{xu2025graphomni}
        & \xmark & \cmark & \xmark & \xmark
        & \xmark & \xmark & \xmark
        & \xmark
        & $5$--$30$
        \\
        GraphInstruct \cite{luo2024graphinstruct}
        & \cmark & \cmark & \xmark & \xmark
        & \xmark & \xmark & \xmark
        & \xmark
        & $5$--$35$
        \\
        GPT4Graph \cite{guo2023gpt4graph}
        & \cmark & \cmark & \cmark & \xmark
        & \cmark & \cmark & \xmark
        & \xmark
        & $10$--$20$
        \\
        NLGraph \cite{wang2023can}
        & \xmark & \cmark & \xmark & \xmark
        & \xmark & \xmark & \xmark
        & \xmark
        & $5$--$35$
        \\
        ProGraph \cite{DBLP:conf/nips/LiCCLSLQW000Y24}
        & \xmark & \cmark & \xmark & \xmark
        & \xmark & \xmark & \xmark
        & \xmark
        & $10$--$10^{6}$
        \\

        LLM4DyG \cite{zhang2024llm4dyg}
        & \cmark & \cmark & \cmark & \xmark
        & \cmark & \xmark & \xmark
        & \xmark
        & $5$--$20$
        \\

        GraphWiz \cite{chen2024graphwiz}
        & \xmark & \cmark & \xmark & \xmark
        & \xmark & \xmark & \xmark
        & \xmark
        & $2$--$100$
        \\

        GraphArena \cite{tang2024grapharena}
        & \xmark & \cmark & \xmark & \xmark
        & \xmark & \cmark & \cmark
        & \xmark
        & $4$--$50$
        \\

        GrAlgoBench \cite{DBLP:journals/corr/abs-2602-06319}
        & \xmark & \cmark & \xmark & \xmark
        & \cmark & \cmark & \xmark
        & \xmark
        & $8$--$160$
        \\

        GraCoRe \cite{yuan2024gracore}
        & \cmark & \cmark & \cmark & \xmark
        & \cmark & \cmark & \xmark
        & \xmark
        & $8$--$30$
        \\
        \toprule

        \textbf{GABench (Ours)}
        & \cmark & \cmark & \cmark & \cmark
        & \cmark & \cmark & \cmark
        & \cmark
        & \textbf{$26$--$3\!\times\!10^{6}$}
        \\

        \bottomrule
    \end{tabularx}
    \vskip  -0.15in
\end{table*}

%% file: preliminary.tex
\subsection{LLM Agents}
% Recently developed Large Language Model (LLM) agents execute complex tasks by integrating reasoning, planning, and tool utilization. 

% An LLM-based agent is defined as an autonomous system that integrates reasoning, planning, and tool utilization powered by llms.
An LLM agent is an autonomous system that employs an LLM to reason, plan, use tools, and interact with an external environment to accomplish user-specified tasks.
Formally, given a user task query $x_q$, a $\theta$-parameterized LLM agent first generates a plan sequence $P = \{p_1, p_2, \dots, p_n\} \sim p_\theta(\cdot \mid x_q)$. Subsequently, the agent executes the corresponding sub-tasks $T = \{t_1, t_2, \dots, t_n\}$ in a sequential manner according to $P$. 
For each sub-task $t_j \in T$, the agent generates the output $y_{t_j}$ according to its associated plan $p_j$, conditioned on the original query $x_q$ and the outputs of all preceding sub-tasks $y_{t_{<j}}$:
$y_{t_j} \sim p_\theta(\cdot \mid x_q, y_{t_{<j}}, p_j)$.
% For each sub-task $t_j \in T$, given its associated plan $p_j$ and the cumulative outputs of all preceding tasks $y_{t_{<j}}$, the agent computes the result $y_{t_j} \sim p_\theta(\cdot \mid x_q, y_{t_{<j}}, p_j)$.
% To solve agentic tasks systematically, agents design execution pipelines known as \textit{agent workflows}. An agent workflow $W = \{w_1, w_2, \dots, w_n\}$ formalizes the task flow by breaking a natural language instruction $u \in \mathcal{U}$ into modular stages $w_i$, each associated with an operation powered by large language models or external tools. 
These interactions can be modeled as a partially observable Markov decision process (POMDP) $(\mathcal{S}, \mathcal{A}, \mathcal{O}, \mathcal{T}, \mathcal{U})$, where $\mathcal{S}$, $\mathcal{A}$, $\mathcal{O}$, and $\mathcal{U}$ denote the state, action, observation, and instruction spaces, respectively. $\mathcal{T}: \mathcal{S} \times \mathcal{A} \rightarrow \mathcal{S} \times \mathcal{O}$ denotes the transition function.
% To characterize the multi-step dynamics of this planning-execution sequence, this agentic task-solving trajectory can be formulated as a partially observable Markov decision process (POMDP) $(\mathcal{S}, \mathcal{A}, \mathcal{O}, \mathcal{T}, \mathcal{U})$ with state space $\mathcal{S}$, action space $\mathcal{A}$, observation space $\mathcal{O}$, transition function $\mathcal{T} : \mathcal{S} \times \mathcal{A} \rightarrow \mathcal{S} \times \mathcal{O}$, and instruction space $\mathcal{U}$.
% To characterize the multi-step dynamics of this planning-execution sequence, this agentic task-solving trajectory can be formulated as a partially observable Markov decision process (POMDP) $(\mathcal{S}, \mathcal{A}, \mathcal{O}, \mathcal{T}, \mathcal{U})$ with state space $\mathcal{S}$, action space $\mathcal{A}$, observation space $\mathcal{O}$, transition function $\mathcal{T} : \mathcal{S} \times \mathcal{A} \rightarrow \mathcal{S} \times \mathcal{O}$, and instruction space $\mathcal{U}$.
Under the POMDP formulation, the interactions can be represented as an agent trajectory $\tau$, which is defined as:
% \begin{equation}
$\tau = (s_0, u, a_0, o_1, a_1, o_2, \dots, a_{T-1}, o_T, s_T)$,
% \end{equation}
where $s_t \in \mathcal{S}$ denotes the environment state at step $t$ (with $s_0$ as the initial state), $u \in \mathcal{U}$ represents the task instruction, $a_t \in \mathcal{A}$ is the action executed by the agent, and $o_t \in \mathcal{O}$ signifies the observation emitted by the environment. This process continues until the agent executes a \texttt{Submit} action or reaches the maximum horizon $T$.

% The iterative interaction in this process between the agent and the environment manifests as a \textit{agent trajectory} $\tau$:
% \begin{equation}
%     \tau = (s_0, u, a_0, o_1, a_1, o_2, \dots, a_{T-1}, o_T, s_T),
% \end{equation}
% where $s_t \in \mathcal{S}$ denotes the environment state at step $t$ (with $s_0$ as the initial state), $u \in \mathcal{U}$ represents the task instruction, $a_t \in \mathcal{A}$ is the action executed by the agent, and $o_t \in \mathcal{O}$ signifies the observation emitted by the environment.
% This interactive process continues until the agent executes a \texttt{Submit} action or reaches the maximum horizon $T$.

\subsection{Graph-Structured Data}
% A graph is defined as $\mathcal{G} = (\mathcal{V}, \mathcal{E}, \mathcal{A}, \mathcal{X})$, where $\mathcal{V}$ represents the set of nodes with cardinality $|\mathcal{V}| = N$, $\mathcal{E}$ represents the set of edges, $\mathcal{A} \in \{0, 1\}^{N \times N}$ denotes the adjacency matrix, and $\mathcal{X}$ encapsulates the associated feature representations of the graph. Depending on the forms of different graph features, $\mathcal{X}$ can be represented in three distinct ways: first, node and edge level textual features can be formalized as $\mathcal{X} = (\mathcal{X}_V, \mathcal{X}_E)$, where each node $v \in \mathcal{V}$ (edge $e \in \mathcal{E}$) corresponds to a raw textual feature $x_v \in \mathcal{X}_V$ ($x_e \in \mathcal{X}_E$); second, graph level textual feature is denoted as $\mathcal{X} = \mathcal{D}$, $\mathcal{D}$ accompanying the entire graph

% A graph can be defined as G = (V, E)，where V = {v1, . . . , v | V | }, E = {e1, . . . , e| E | } are the node set, edge set, , respectively. The adjacency matrix of the graph G is denoted as A ∈ R| V |×| V | , where Ai j = 1 if nodes vi and v j are connected, otherwise Ai j = 0.
% (1)Numerical-Attribute Graph  XXX
% (2)Text-Attributed Graphs (TAG) can be denoted as G = (V, E, T)，where T XXX...Each vi ∈ V is associated with some textual information tvi ∈ T. 
% (3)Text-Paired Graph can be denoted as the pair (G, dG), where dG  represents the textual description of the entire graph..
A graph is denoted by $\mathcal{G}=(\mathcal{V},\mathcal{E})$, where $\mathcal{V}=\{v_1,\ldots,v_{|\mathcal{V}|}\}$ and $\mathcal{E}=\{e_1,\ldots,e_{|\mathcal{E}|}\}$ denote the node and edge sets, respectively. The adjacency matrix of the graph $\mathcal{G}$ is denoted as $\mathbf{A} \in \mathbb{R}^{|\mathcal{V}| \times |\mathcal{V}|}$, where $\mathbf{A}_{ij}=1$ if nodes $v_i$ and $v_j$ are connected and $A_{ij}=0$ otherwise.
In this work, we consider three common graph types with different forms of attribute information.
(1) \textbf{Numerical-Attribute Graphs (NAGs)} with node-level numerical attributes are denoted by
$\mathcal{G}=(\mathcal{V},\mathcal{E},\mathbf{X})$, where $\mathbf{X}\in\mathbb{R}^{|\mathcal{V}|\times d}$ is the $d$-dimensional node attribute matrix and each row $\mathbf{x}_i$ represents the numerical attributes of node $v_i$.
(2) \textbf{Text-Attributed Graphs (TAGs)} with node-level textual attributes are denoted by
$\mathcal{G}=(\mathcal{V},\mathcal{E},\mathcal{T})$, where $\mathcal{T}=\{t_1,\ldots,t_{|\mathcal{V}|}\}$ is the set of texts, and each node $v_i$ is associated with node-level text $t_i$.
(3) \textbf{Text-Paired Graphs (TPGs)} with graph-level textual information are represented as graph-text pairs
$(\mathcal{G},d_{\mathcal{G}})$, where $d_{\mathcal{G}}$ is a textual description associated with the entire graph.

\begin{figure*}[t]
  \centering
  \vskip -0.05in
  \includegraphics[width=0.97\textwidth]{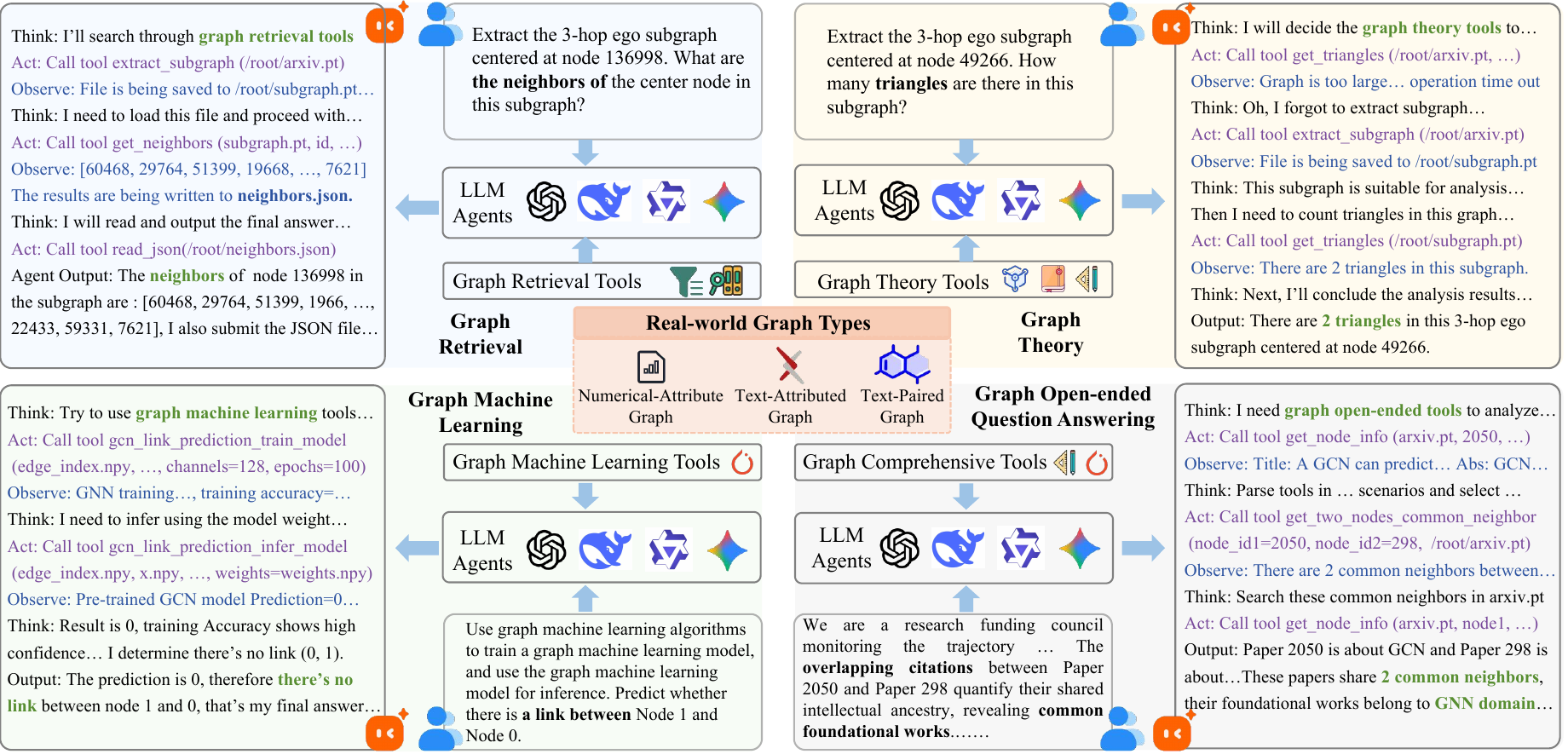}
  \vskip -0.15in
  \caption{Overview of GABench. The benchmark evaluates LLM agents across four agentic graph analysis categories: graph retrieval, graph theory, graph machine learning, and graph open-ended question answering. 
  GABench also provides an executable graph analysis toolset covering all task categories. We show a concrete example for each category of agentic tasks.}
  \vskip -0.15in
  \label{fig:GABench}
\end{figure*}
% \subsection{Graph Analysis Tasks}
\subsection{Agent-Based Graph Analysis}
Given a graph $\mathcal{G}$, a graph analysis task involves analyzing 
its topology and attributes to retrieve graph information, compute 
structural properties, make graph learning predictions, or derive 
higher-level insights. We denote the ground-truth result of the task 
by $y^{*}$. In the agent setting, the graph $\mathcal{G}$ is stored in an external  environment rather than being directly provided in the task instruction $u$. The agent follows the trajectory $\tau$ defined above, invoking graph tools and using the resulting observations to produce a final answer $\hat{y}$. The task is successfully completed when $\hat{y}$ satisfies the instruction $u$ and is consistent with the ground-truth result $y^{*}$.

%% file: benchmark.tex
\subsection{Overview and Statistics}
To comprehensively evaluate the planning, tool-use, and end-to-end problem-solving capabilities of LLM agents on realistic graph analysis tasks, we propose GABench. 
As illustrated in Figure~\ref{fig:GABench}, GABench integrates multiple graph types, graph analysis tasks, executable tools, and agent--tool interactions within a unified evaluation framework.
% GABench is designed to evaluate the planning, tool-use, and end-to-end problem-solving capabilities of LLM agents on realistic graph analysis tasks. To support a comprehensive evaluation, 
Specifically, GABench collects 13 real-world datasets from 6 domains and covers 3 different graph types: NAGs, TAGs, and TPGs. These datasets range from 26 to more than 3.7 million nodes and contain up to 123.6 million edges.
GABench further covers 4 common graph analysis categories: graph retrieval for accessing explicit graph information; graph theory for computing structural properties; graph machine learning for node classification, link prediction, and graph classification; and graph open-ended question answering for graph insight and prediction. Based on these datasets and tasks, GABench provides a unified toolset of 84 executable tools, including 10 for graph retrieval, 54 for graph theory, 10 for graph machine learning, and 10 for open-ended graph question answering.
Finally, GABench introduces an agentic graph task generation pipeline that transforms existing graph data and tasks into executable and verifiable agentic tasks. 
Each task consists of a graph analysis question, a ground-truth tool trajectory, and a corresponding answer. 
Table~\ref{tab:graph_tblr} summarizes the 10,400 evaluation instances in GABench.

\input{benchmark_statistics}

\subsection{Raw Data Collection}
To ensure a comprehensive evaluation across diverse real-world applications, we consider four key criteria when selecting datasets for GABench: (1) \textbf{Various graph types.} We consider three common graph types based on attribute forms, including NAGs, TAGs, and TPGs. 
(2) \textbf{Various domains.} 
Datasets in GABench span multiple domains, including e-commerce, citation networks, chemistry, finance, social networks, and justice.
(3) \textbf{Diverse scales and densities.} We consider a wide range of graph scales, spanning from tens to millions of nodes ($26$ to over $3$ million). 
(4) \textbf{Tool compatibility and task coverage}. Each dataset must support executable operations across multiple graph analysis tasks and provide sufficient information for deterministic answer construction.
% Such diversity in graph types, domains, and scales, enables a comprehensive assessment of LLM agents across a wide range of graph analysis contexts. 

Based on these criteria, GABench collects 13 datasets from 6 domains, as summarized in Table~\ref{tab:graph_statistics} and Table~\ref{tab:benchmark_comparison}. These datasets are categorized into 3 graph types: 
% (i) Text-attributed graphs. We only consider text-attributed graphs where each node represents a textual entity and is associated with a textual description. (ii) Various domains. Datasets in GLBench span multiple domains, including citation networks, web links, and social networks. (iii) Diverse scale and density. GLBench datasets cover a wide range of scales, from thousands to hundreds of thousands of nodes. The density also varies significantly, with average node degrees ranging from 2.6 (i.e., Citeseer) to 36.9 (i.e., WikiCS).
% \textbf{(1) Numerical-Attribute Graphs (NAGs).} 
(1) For NAGs with node-level numerical attributes, we choose Pokec and NBA from the social domain, DGraph-Fin and Credit from the finance domain, and Bail from the justice domain.
%In this category, nodes are associated with node-level numerical attributes. These datasets span three domains: social, finance, and justice. Specifically, the social domain comprises Pokec and NBA. The finance domain includes DGraph-Fin and Credit, while the justice domain contains Bail.
% \textbf{(2) Text-Attributed Graphs (TAGs).} 
(2) For TAGs with node-level textual attributes, we choose Arxiv and Cora from the academic citation domain, Reddit and Instagram from the social domain, History and Product from the e-commerce domain.
%In this category, each node is associated with node-level textual attributes. The selected TAGs span three domains: academic citation, social networks, and e-commerce. Specifically, the academic citation domain includes arXiv and Cora. The social domain comprises Reddit and Instagram. The e-commerce domain consists of History and Product.
% \textbf{(3) Text-Paired Graphs (TPGs).} 
(3) For TPGs with graph-level textual information, we choose HIV and PCBA from the molecular domain. 
Further details about these datasets are provided in Appendix~\ref{app:datasets}.
%In this category, each graph is paired with a graph-level textual description. Our benchmark focuses on the molecule domain, including HIV and PCBA.
% \subsection{Agentic Graph Tasks}
\subsection{Task Selection}
% Graph Retrieval
% Graph Theory
% Graph Machine Learning
% Open-ended Graph Question Answering

To comprehensively cover real-world graph tasks, GABench considers 4 common categories:
% To comprehensive cover reale-world graph analysis task, GABench considers 4 common analysis senarios：
% To cover the primary objectives of practical graph analysis, GABench considers 4 common task: 
graph retrieval, graph theory, graph machine learning, and graph open-ended question answering.

% Definition → Practical role → Included subtasks → Expected outputs → Agent capabilities evaluated
\textbf{Graph retrieval} tasks aim to access explicit information stored in a graph, such as node attributes, local neighborhoods, and graph statistics. GABench includes node-, edge-, and graph-level retrieval tasks. Node-level tasks query the attributes, neighbors, or degrees of specified nodes. Edge-level tasks query whether an edge exists between two nodes or retrieve its associated information. Graph-level tasks retrieve graph statistics or extract a specified subgraph. Given the target graph objects and retrieval conditions, these tasks return identifiers, attributes, Boolean values, numerical statistics, or subgraphs. They evaluate whether an agent can understand the requested information, locate the relevant graph objects, and use appropriate graph-access tools.

\textbf{Graph theory} tasks aim to compute structural properties using deterministic graph algorithms. 
% These tasks characterize the roles of nodes and edges and the overall graph structure. 
Here, GABench includes node-, edge-, and graph-level graph theory tasks. Specifically, node-level tasks compute properties of individual nodes, such as centrality and clustering coefficients. Edge-level tasks analyze relations between nodes or edges, such as shortest paths, bridges, and node similarity. Graph-level tasks compute global properties, such as cycles and graph diameter. Outputs may be numerical values, paths, sets of nodes or edges, or Boolean results. These tasks evaluate whether an agent can select a suitable algorithm, provide correct inputs, execute the tool, and return the result.

\textbf{Graph machine learning} tasks aim to predict unknown labels, links, or graph properties using graph topology, attributes, and available supervision. GABench includes 3 common prediction tasks: node classification, link prediction, and graph classification. Node classification predicts the label of a target node, link prediction estimates whether a relation exists between a pair of nodes, and graph classification predicts a label or property of an entire graph. Given graph data, prediction targets, and the required graph machine learning inputs, these tasks return predicted labels or scores. They evaluate whether an agent can prepare the required graph data, select and configure the provided graph learning tools, and obtain the corresponding prediction results.

\textbf{Graph open-ended question answering} tasks aim to answer scenario-based questions with free-form responses. Unlike tasks with fixed output formats, these questions require the agent to determine what graph information or analysis is needed according to the user’s request. GABench includes graph insight and graph prediction tasks. Graph insight tasks ask the agent to identify and explain patterns, relationships, or structural characteristics in a graph. Graph prediction tasks ask the agent to make and explain predictions about graph entities or relations. Their outputs are textual answers that should directly address the question and remain consistent with the underlying graph evidence. These tasks evaluate whether an agent can understand a graph analysis request, select suitable operations, and present the findings clearly and accurately.

%Graph open-ended question answering tasks aim to answer scenario-based questions with free-form responses grounded in graph data. Unlike tasks with fixed output formats, these questions require the agent to determine what graph information or analysis is needed according to the user request. GABench includes graph insight and graph prediction tasks. Graph insight tasks ask the agent to identify and explain patterns, relationships, or structural characteristics in a graph. Graph prediction tasks ask the agent to make and explain predictions about graph entities or relations. Their outputs are textual answers that should directly address the question and remain consistent with the underlying graph evidence. These tasks evaluate whether an agent can understand a high-level graph analysis request, select suitable analysis operations, and present the obtained findings in a clear and accurate response.
\begin{figure*}[t]
  \centering
  \includegraphics[width=1.0\textwidth]{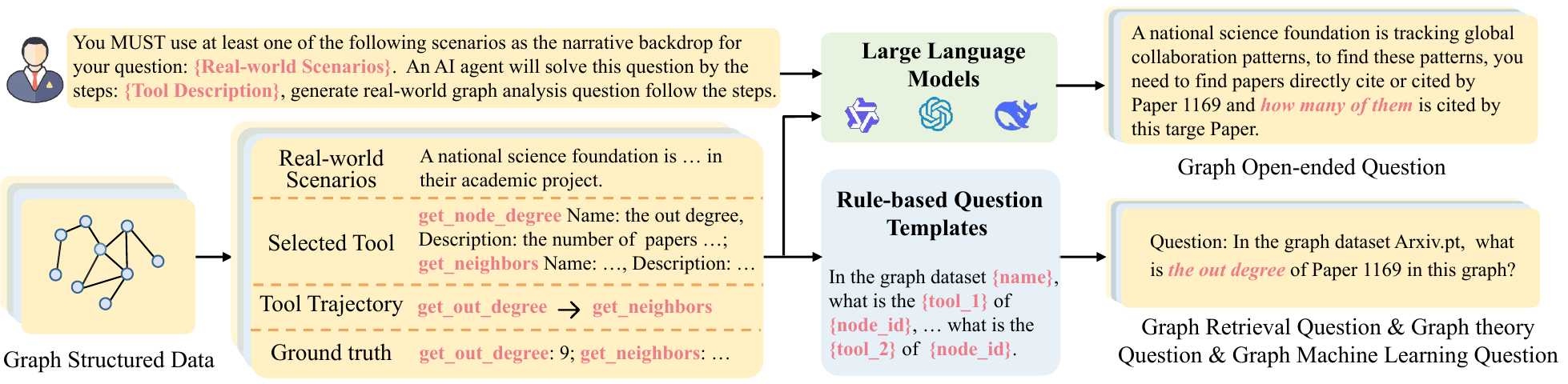}
  \vskip -0.1in
  \caption{LLM-based pipeline for generating open-ended graph question-answering tasks.}
  \vskip -0.15in
  \label{fig:pipeline}
\end{figure*}

\subsection{Graph Analysis Tool Set}
Here, GABench provides a unified toolset that covers graph data access and four categories of graph tasks. 
% Before performing task-specific operations, agents can inspect graph files, extract required attributes or individual graphs, and convert the extracted data into formats used by subsequent analysis tools. 
As summarized in Table~\ref{tab:graph_tblr}, the task-specific toolset contains 84 tools, including 10 for graph retrieval, 54 for graph theory, 10 for graph machine learning, and 10 for open-ended graph question answering.
Specifically, the graph retrieval tools allow agents to query node attributes, check edge existence, extract subgraphs, and count nodes or edges. The graph theory tools cover both local and global structural analysis, including centrality and clustering coefficients, graph traversal, connectivity, and cycle detection. 
% These tools provide node-, edge-, and graph-level operations for NAGs, TAGs, and TPGs. 
The graph machine learning tools support node classification and link prediction on NAGs and TAGs, as well as graph classification on TPGs. To reduce execution cost and improve evaluation stability, they emulate standard GNN training by validating input parameters and returning a fixed accuracy for all valid tasks without actual model training, following prior studies~\cite{qin2024nasbenchgraphbenchmarkinggraphneural,yang2024cragcomprehensiverag,ruan2024identifyingriskslmagents}. The graph open-ended question answering tools support graph insight and graph prediction by combining global and local operations, such as shortest-path analysis and community detection. Together, these tools provide the operations required by the four task categories across the three graph types.

% \subsection{Graph Analysis Task Generation Pipeline}
\subsection{Agentic Graph Task Generation Pipeline}
% in this
In this subsection, we present the overall design of our agentic graph task generation pipeline, including the construction of executable tool trajectories, verifiable answers, and corresponding user questions.
%In practice, constructing agentic graph tasks typically involves executable graph tools, verifiable answers, and graph analysis questions.
To meet these requirements, we develop an execution-grounded pipeline built on a unified graph analysis toolset, as illustrated in Figure~\ref{fig:pipeline}. 
For each target task, the pipeline first selects and executes the required tools to construct a ground-truth tool trajectory and answer, and then synthesizes the corresponding user question from this trajectory. Questions for graph retrieval, graph theory, and graph machine learning are generated using rule-based templates. However, since graph open-ended questions must reflect real-world application contexts, we use an LLM-based process that combines realistic scenarios and tool descriptions to generate them.

\paragraph{\textbf{Tool Trajectory and Answer Construction}}
% To ensure that each task has an executable solution path and a verifiable answer, we construct and execute a ground-truth tool trajectory for every task. Given a graph and a task, we select the required tools and organize them into stages according to their input--output dependencies. 
% We assign each selected tool a stage index according by:
To ensure that every task has an executable solution path and a verifiable answer, we construct and execute a ground-truth tool trajectory. Given a graph and a task, we select the required tools and assign them to stages based on their input--output dependencies:
\begin{equation}
  s(t)=
  \begin{cases}
  1, & \mathcal{P}(t)=\emptyset,\\[3pt]
  1+\max\limits_{u\in\mathcal{P}(t)} s(u), & \text{otherwise},
\end{cases}
\end{equation}
where $\mathcal{P}(t)$ denotes the set of direct predecessors of tool $t$, $u\in\mathcal{P}(t)$ denotes any such predecessor, and $s(t)$ denotes the stage index of $t$.
For each stage, we randomly select one or more tools and execute the selected tools in ascending stage order to construct the ground-truth tool trajectory,  with their execution results used to derive the ground-truth answer.

% where $t$ is the target tool and $u$ is any direct predecessor tool of $t$.
% where $t$ denotes the tool whose stage is being determined, and $u$ denotes any direct predecessor tool of $t$. 
% The function $\mathcal{P}(\cdot)$ maps each tool to its direct predecessors. $\mathcal{P}(t)$ is the tool set whose outputs as the inputs of the tool $t$. The stage function $s(\cdot)$ maps each tool to a positive integer stage index: $s(t)$ is the stage index assigned to $t$, while $s(u)$ is the stage index assigned to predecessor $u$.

\paragraph{\textbf{Graph Analysis Question Generation.}}
Next, we generate questions based on the constructed tool trajectories, tool names and answers, as illustrated in Figure~\ref{fig:pipeline}. For graph retrieval, graph theory, and graph machine learning, we use rule-based templates, such as ``In the graph dataset \{graph\_name\}, what is the \{tool\_name\} of \{node\_id\}?''. For graph open-ended questions that commonly arise in real-world applications, we propose an LLM-based question generation pipeline, which provides the LLM with real-world scenarios and descriptions of the selected tools. Specifically, to generate real-world scenarios, we query the LLM with a type-specific graph description. We use GPT-4o-mini for all LLM-based question generation steps.
%graph open-ended question answering usually requires questions that reflect the background of a specific application domain. We therefore propose an LLM-based pipeline to generate natural user requests for open-ended graph analysis by providing the LLM with the scenarios and descriptions of the selected tools.
For TAGs, we extract a subgraph and describe its node texts and edge relations in natural language. For NAGs and TPGs, we instead provide a graph-level summary, such as ``This is a molecular graph in which each node represents an atom and each edge represents a chemical bond.'' The LLM then generates a realistic scenario using the following prompt:
% \vskip -0.1in
\begin{tcolorbox}[before skip=3pt, after skip=3pt, top=1pt, bottom=1pt, left=2pt, right=2pt]
\textbf{System prompt:} You are an expert in designing high-level real-world scenarios... for graph datasets.\\
\textbf{Rules:} Domain-Level: Focus on broad domain background without including specific node/edge entities or names.\\
\textbf{User content:} \{graph\_context\}. Please generate 5 to 10 abstract, high-level real-world background scenarios for this domain. 
\end{tcolorbox}
In practice, users typically express their objectives without specifying tool names or invocation formats. Therefore, for each selected tool, we use the LLM to generate a description of its function in the given graph scenario, using the following prompt:
\begin{tcolorbox}[before skip=3pt, after skip=3pt, top=1pt, bottom=1pt, left=2pt, right=2pt]
\textbf{System prompt:} Your task is to rewrite generic graph-theory tool descriptions... into domain-specific descriptions. \\
\textbf{Rules:} 1. CRITICAL: Do NOT use... 2. Do not use any algorithm names or graph-theory terminology.\\
\textbf{User content:}
Use the following overview... Graph Overview:
\{TAG\_subgraph\_description\}
\{graph\_level\_Summary\} \\
\textbf{Tools to Rewrite:}
\{tool\_list\}. Generate domain-adaptive, fuzzy descriptions for the listed graph tools.
\end{tcolorbox}

\input{retrieval_exp}

\input{theory_exp}
Finally, we provide the LLM with the scenario and descriptions of the tools used in the ground-truth trajectory. The LLM then generates a natural user question that situates the graph analysis task within this scenario using the following prompt. We then package the generated question, ground-truth tool trajectory, and corresponding answer into a benchmark instance for evaluation. The complete prompt is provided in Appendix~\ref{app:generation_prompt}.
%Finally, we provide the LLM with the generated scenario and descriptions of the tools used in the ground-truth trajectory. The LLM then generates a natural user question that situates the graph analysis task within the generated graph scenario, using the following prompt. After completing this process, we package the generated question, the ground-truth tool trajectory, and the corresponding answer into a benchmark instance for evaluation.
%The complete prompt is provided in Appendix~\ref{app:generation_prompt}.
\begin{tcolorbox}[before skip=3pt, after skip=2pt, top=1pt, bottom=1pt, left=2pt, right=2pt]
\textbf{System prompt:} Write one graph-analysis question..., based on a pre-defined sequence of graph-analysis steps... \\
\textbf{Rules:} Cascading Storytelling: Weave the analytical steps into a causal narrative where each step naturally sets up the next. \\
\textbf{User content:} Follow the Rules: Graph Scenarios in the domain "\{graph\_scenarios\}". Tool Descriptions: "\{tool\_descriptions\}"
\end{tcolorbox}

\paragraph{\textbf{Task Validation and Quality Control}}
We validate each generated task through execution checks, question-level checks, and human review. First, we verify that the selected tools are applicable to the target graph and task and that all tool calls execute successfully within a predefined time limit. We then ensure that the generated question preserves the intended requirements, can be answered using the recorded ground-truth trajectory, and does not reveal tool names, intermediate outputs, or the final answer. Finally, human annotators assess clarity, realism, domain consistency, task-category correctness, and overall validity. Tasks with invalid or timed-out tool calls, ambiguous questions, graph-data inconsistencies, or insufficient tool support are removed.

% We validate each generated task at both the execution and language levels. During tool selection and ground-truth generation, we assess whether each selected tool is appropriate for the target graph and task. A task is discarded if any tool call is invalid, fails to execute, or exceeds a predefined time limit. We further check whether the generated question preserves the original task conditions, is fully answerable using the recorded trajectory, and does not expose tool names, intermediate results, or the ground-truth answer. Finally, human annotators inspect the generated tasks for clarity, realism, domain consistency, task-category correctness, and scientific validity. Tasks that are ambiguous, inconsistent with the graph data, or unsupported by the provided tools are revised or removed.

\subsection{Evaluation Metrics}
\paragraph{\textbf{Rule-based Evaluation}} 
To evaluate whether LLM agents select the required tools in the correct order, we use tool selection accuracy \textbf{(TSA)}~\cite{ruan2024identifyingriskslmagents}. 
% TSA measures the alignment between the tool sequence executed by an agent and the corresponding ground-truth tool trajectory.
% Let $S_{\mathrm{agent}}^{(i)}$ denote the sequence of tools invoked by the agent for the $i$-th task, and $S_{\mathrm{true}}^{(i)}$ denote the corresponding ground-truth sequence. 
Let $S_{\mathrm{agent}}^{(i)}$ and $S_{\mathrm{true}}^{(i)}$ denote the agent-generated and ground-truth tool sequences for task $i$, respectively.
The longest common subsequence (LCS) is the longest ordered subsequence shared by two sequences.
We compute the LCS between $S_{\mathrm{agent}}^{(i)}$ and $S_{\mathrm{true}}^{(i)}$, which represents the longest correctly ordered sequence of selected tools.
For a dataset containing $N$ tasks, TSA is defined as $\mathrm{TSA}\!=\!\sum_{i=1}^{N}|\mathrm{LCS}(S_{\mathrm{agent}}^{(i)},S_{\mathrm{true}}^{(i)})|\Big/\sum_{i=1}^{N}|S_{\mathrm{true}}^{(i)}|$, 
where $|\cdot|$ denotes the sequence length.
A higher TSA indicates a more complete and correctly ordered tool-selection sequence.

\paragraph{\textbf{LLM-as-a-Judge Evaluation}.}
We use an LLM-as-a-Judge to assess each agent's final response and compute the task success rate \textbf{(SR)}~\cite{ma2024agentboard}. This evaluator supports semantic comparison of free-form responses, which may contain diverse answer formats and intermediate reasoning. Specifically, DeepSeek-V4-Flash receives the original question, the ground-truth answer, and the agent's final response, and assigns a binary score based on their semantic consistency. A score of $1$ indicates successful task completion, whereas $0$ indicates failure. The detailed evaluation protocol is provided in Appendix~\ref{app:judge_protocol}, and the prompt is provided in Appendix~\ref{app:judge_prompt}.
%We use an LLM-as-a-Judge to assess the correctness of each agent's final response and compute the task success rate \textbf{(SR)}~\cite{ma2024agentboard}. An LLM-based evaluator is necessary for two reasons. First, responses to graph open-ended question-answering tasks may combine multiple graph findings and explanations in diverse formats, making semantically equivalent answers difficult to evaluate through fixed-format or exact-match rules. Second, even for graph retrieval, graph theory, and graph machine learning tasks with verifiable answers, agents often produce reasoning-rich responses that contain intermediate hypotheses, tentative conclusions, and self-corrections before presenting the final answer. Rule-based extraction can therefore be unreliable, as it may mistake an intermediate result for the agent's final conclusion or fail to identify the answer expressed in free-form text. Specifically, we employ DeepSeek-V4-Flash as the judge model. For each task, the judge receives the original question, the ground-truth answer, and the final response produced by the agent. It compares the agent response with the ground truth based on semantic consistency and returns a binary score. A score of $1$ indicates successful task completion, whereas a score of $0$ indicates failure. 
%The evaluation protocol and judge prompt are provided in Appendix~\ref{app:judge_prompt}.

%% file: benchmark_statistics.tex
\begin{table*}[t]
\centering
\vskip -0.05in
\caption{Composition of GABench by task category and graph type.}
\vskip -0.15in
\label{tab:graph_tblr}
\small
\begin{tblr}{
  width = \textwidth,
  colspec = {
    Q[c,m,wd=2.2cm]
    Q[c,m,wd=1.1cm]
    Q[c,m,wd=1.8cm]
    Q[c,m,wd=1.4cm]
    Q[c,m,wd=2.2cm]
    X[c,m]
  },
  colsep = 2pt,
  cells = {valign=m},
  column{1-3} = {leftsep=-2pt, rightsep=-2pt},
  row{1} = {font=\bfseries, halign=c, valign=m, abovesep=4pt, belowsep=4pt},
}
\toprule
\SetCell{halign=c,valign=m} \begin{tabular}[c]{@{}c@{}}Task\\Category\end{tabular}
& \SetCell{c,m} \#Tools
& \SetCell{halign=c,valign=m} \begin{tabular}[c]{@{}c@{}}Average\\ \#Tool Calls\end{tabular}
& \SetCell{c,m} Graph Type
& \SetCell{c,m} Task
& \SetCell{c,m} Source Dataset (\#Samples)
\\
\midrule

{}
& \SetCell[r=3]{c,m} 10
& \SetCell[r=3]{c,m} 3
& NAG
& node, edge, graph
& DGraph-Fin (200), Pokec (200), NBA (200), Credit (200), Bail (200)
\\
\cmidrule[lr]{4-6}

\smash{Graph Retrieval} & & &
TAG
& node, edge, graph
& Reddit (200), Instagram (200), History (200), Product (200), Arxiv (200), Cora (200)
\\
\cmidrule[lr]{4-6}

& & &
TPG
& node, edge, graph
& HIV (200), PCBA (200)
\\

\midrule

{}
& \SetCell[r=3]{c,m} 54
& \SetCell[r=3]{c,m} 3
& NAG
& node, edge, graph
& DGraph-Fin (200), Pokec (200), NBA (200), Credit (200), Bail (200)
\\
\cmidrule[lr]{4-6}

\smash{Graph Theory} & & &
TAG
& node, edge, graph
& Reddit (200), Instagram (200), History (200), Product (200), Arxiv (200), Cora (200)
\\
\cmidrule[lr]{4-6}

& & &
TPG
& node, edge, graph
& HIV (200), PCBA (200)
\\

\midrule

{}
& \SetCell[r=3]{c,m} 10
& \SetCell[r=3]{c,m} 4
& NAG
& NC, LP
& DGraph-Fin (200), Pokec (200), NBA (200), Credit (200), Bail (200)
\\
\cmidrule[lr]{4-6}

\smash{\raisebox{-0.5\height}{\shortstack{Graph Machine\\Learning}}} & & &
TAG
& NC, LP
& Reddit (200), Instagram (200), History (200), Product (200), Arxiv (200), Cora (200)
\\
\cmidrule[lr]{4-6}

& & &
TPG
& GC
& HIV (200), PCBA (200)
\\

\midrule

{}
& \SetCell[r=3]{c,m} 10
& \SetCell[r=3]{c,m} 7
& NAG
& insight, prediction
& DGraph-Fin (200), Pokec (200), NBA (200), Credit (200), Bail (200)
\\
\cmidrule[lr]{4-6}

\smash{\raisebox{-0.5\height}{\shortstack{Graph Open-ended\\Question \\Answering}}} & & &
TAG
& insight, prediction
& Reddit (200), Instagram (200), History (200), Product (200), Arxiv (200), Cora (200)
\\
\cmidrule[lr]{4-6}

& & &
TPG
& insight, prediction
& HIV (200), PCBA (200)
\\

\bottomrule
\end{tblr}
\vskip -0.15in
\end{table*}

%% file: retrieval_exp.tex
\begin{table*}[t]
\centering
\caption{LLM-agent performance on graph retrieval tasks. The best results are \textbf{bolded} and the second-best results are \underline{underlined}.}
\label{tab:retrieval_exp}
\vskip -0.15in
\resizebox{\textwidth}{!}{%
\setlength{\tabcolsep}{3pt}%
\begin{tabular}{l *{18}{c}}
\toprule
\multirow{3}{*}[-2pt]{\textbf{Model}}
& \multicolumn{6}{c}{\textbf{Numerical-Attribute Graph}}
& \multicolumn{6}{c}{\textbf{Text-Attributed Graph}}
& \multicolumn{6}{c}{\textbf{Text-Paired Graph}} \\[-2pt]
\cmidrule(lr){2-7}
\cmidrule(lr){8-13}
\cmidrule(lr){14-19}

& \multicolumn{2}{c}{Node}
& \multicolumn{2}{c}{Edge}
& \multicolumn{2}{c}{Graph}
& \multicolumn{2}{c}{Node}
& \multicolumn{2}{c}{Edge}
& \multicolumn{2}{c}{Graph}
& \multicolumn{2}{c}{Node}
& \multicolumn{2}{c}{Edge}
& \multicolumn{2}{c}{Graph} \\[-2pt]
\cmidrule(lr){2-3}
\cmidrule(lr){4-5}
\cmidrule(lr){6-7}
\cmidrule(lr){8-9}
\cmidrule(lr){10-11}
\cmidrule(lr){12-13}
\cmidrule(lr){14-15}
\cmidrule(lr){16-17}
\cmidrule(lr){18-19}

& SR & TSA
& SR & TSA
& SR & TSA
& SR & TSA
& SR & TSA
& SR & TSA
& SR & TSA
& SR & TSA
& SR & TSA \\[-1pt]

\midrule

Qwen3-235B
& 7.50 & 5.13
& 6.66 & 6.67
& 5.00 & 5.90
& 22.62 & 6.25
& 6.94 & 9.97
& 5.48 & 5.56
& 0.43 & 1.86
& 25.00 & 7.14
& 1.62 & 5.26 \\[1pt]

GLM-4.5-flash
& 1.38 & 37.50
& 4.73 & 41.83
& 1.62 & 35.72
& 48.78 & 73.04
& 32.46 & 48.39
& 30.63 & 56.04
& 28.57 & \underline{57.27}
& 7.12 & 54.24
& 7.14 & 48.86 \\[1pt]

GLM-5-turbo
& 6.66 & 63.23
& 5.00 & 67.19
& \underline{5.34} & 64.71
& 60.76 & 59.01
& 15.15 & 69.92
& 22.73 & 54.33
& \underline{60.00} & 55.89
& 11.11 & 57.69
& 42.86 & 47.37 \\[1pt]

DeepSeek-V4-pro
& \textbf{24.82} & \underline{63.64}
& \underline{59.66} & \underline{71.97}
& \textbf{8.27} & \underline{67.06}
& \underline{62.26} & \textbf{98.29}
& \textbf{79.68} & \textbf{99.24}
& \textbf{71.97} & \textbf{86.71}
& 45.46 & 42.42
& \textbf{68.00} & \underline{74.79}
& \textbf{90.39} & \textbf{70.74} \\[1pt]

GPT-5-mini
& 12.46 & 23.23
& \textbf{70.59} & 28.28
& 4.00 & 30.86
& \textbf{63.04} & 25.00
& \underline{78.17} & 28.79
& \underline{66.05} & 28.33
& 50.00 & 26.52
& 56.37 & 21.97
& \underline{45.61} & 35.18 \\[1pt]

Gemini-3-flash
& \underline{20.65} & \textbf{77.94}
& 0.80 & \textbf{85.93}
& \underline{5.34} & \textbf{80.01}
& 40.20 & \underline{81.52}
& 5.00 & \underline{90.49}
& 15.63 & \underline{71.80}
& \textbf{70.59} & \textbf{80.87}
& \underline{62.50} & \textbf{81.25}
& 12.12 & \underline{69.64} \\[1pt]

\bottomrule
\end{tabular}%
}
% \vskip -0.05in
\end{table*}

%% file: theory_exp.tex
\begin{table*}[t]
\centering
\caption{LLM-agent performance on graph theory tasks. The best results are \textbf{bolded} and the second-best results are \underline{underlined}.}
\label{tab:theory_exp}
\vskip -0.15in
\resizebox{\textwidth}{!}{%
\setlength{\tabcolsep}{2.5pt}%
\begin{tabular}{l *{18}{c}}
\toprule
\multirow{3}{*}[-2pt]{\textbf{Model}}
& \multicolumn{6}{c}{\textbf{Numerical-Attribute Graph}}
& \multicolumn{6}{c}{\textbf{Text-Attributed Graph}}
& \multicolumn{6}{c}{\textbf{Text-Paired Graph}} \\[-2pt]
\cmidrule(lr){2-7}
\cmidrule(lr){8-13}
\cmidrule(lr){14-19}

& \multicolumn{2}{c}{Node}
& \multicolumn{2}{c}{Edge}
& \multicolumn{2}{c}{Graph}
& \multicolumn{2}{c}{Node}
& \multicolumn{2}{c}{Edge}
& \multicolumn{2}{c}{Graph}
& \multicolumn{2}{c}{Node}
& \multicolumn{2}{c}{Edge}
& \multicolumn{2}{c}{Graph} \\[-2pt]
\cmidrule(lr){2-3}
\cmidrule(lr){4-5}
\cmidrule(lr){6-7}
\cmidrule(lr){8-9}
\cmidrule(lr){10-11}
\cmidrule(lr){12-13}
\cmidrule(lr){14-15}
\cmidrule(lr){16-17}
\cmidrule(lr){18-19}

& SR & TSA & SR & TSA & SR & TSA
& SR & TSA & SR & TSA & SR & TSA
& SR & TSA & SR & TSA & SR & TSA \\[-1pt]
\midrule

Qwen3-235B
& 12.90 & 6.16
& 6.15 & 3.57
& 1.82 & 5.62
& 25.54 & 12.46
& 8.33 & 9.97
& 13.57 & 5.55
& 15.00 & 9.48
& 10.00 & 7.14
& 7.14 & 5.26 \\[1pt]

GLM-4.5-flash
& 8.00 & 35.90
& 8.00 & 21.53
& 8.89 & 36.67
& 53.16 & 65.52
& 47.92 & 48.20
& 54.08 & 39.83
& 17.39 & 50.55
& 16.67 & 42.71
& 13.57 & 43.50 \\[1pt]

GLM-5-turbo
& 19.20 & 47.06
& \underline{21.45} & \underline{54.69}
& 5.92 & \underline{55.00}
& 50.85 & 54.12
& 7.93 & 54.35
& 18.05 & 52.00
& 28.00 & 38.23
& 13.33 & 39.07
& 10.49 & 34.99 \\[1pt]

DeepSeek-V4-pro
& \underline{20.22} & \underline{57.58}
& \textbf{28.37} & 53.79
& \textbf{29.10} & 52.40
& \textbf{64.63} & \textbf{98.10}
& \underline{54.65} & \textbf{98.11}
& \textbf{73.99} & \textbf{98.53}
& \underline{64.29} & \underline{57.57}
& \textbf{62.15} & \underline{63.24}
& \textbf{78.70} & \underline{76.47} \\[1pt]

GPT-5-mini
& \textbf{28.91} & 23.23
& 20.21 & 23.23
& \underline{22.54} & 30.85
& \underline{58.14} & 25.82
& \textbf{66.03} & 26.89
& \underline{64.50} & 28.33
& 49.12 & 26.51
& \underline{48.39} & 21.97
& \underline{46.43} & 70.36 \\[1pt]

Gemini-3-flash
& 18.06 & \textbf{72.06}
& 0.80 & \textbf{71.87}
& 5.34 & \textbf{72.05}
& 40.20 & \underline{83.96}
& 5.10 & \underline{85.62}
& 15.63 & \underline{78.57}
& \textbf{70.58} & \textbf{70.59}
& 6.25 & \textbf{65.62}
& 12.12 & \textbf{86.77} \\[1pt]

\bottomrule
\end{tabular}%
}
\vskip -0.05in
\end{table*}

%% file: experiment.tex
\subsection{Experiment Settings}
%We select OpenClaw as our primary evaluation harness because it provides a production-grade CLI environment with direct, containerized access to real system tools such as shells and file systems. To investigate the performance of agent systems under complex and long-horizon workflows, we need to select frontier models that represent the current state of the art.
\paragraph{\textbf{Models and Harnesses}} 
OpenClaw~\cite{openclaw} provides containerized access to system tools, including shells and file systems, and serves as our primary evaluation harness. Under OpenClaw, we evaluate 6 frontier LLMs on GABench: Qwen3-235B~\cite{yang2025qwen3}, GLM-4.5-flash~\cite{zeng2025glm}, GLM-5-turbo~\cite{glm5team2026glm5}, DeepSeek-V4-pro~\cite{deepseekai2026deepseekv4}, GPT-5-mini~\cite{openai2025gpt5}, and Gemini-3-flash~\cite{geminiteam2025gemini}.
% We also examine the effect of different agent harnesses by evaluating the same model, GLM-5-turbo, with OpenClaw, Claude Code, and Hermes Agent. 
% For model comparisons within the same harness, we use identical tool schemas, system prompts, and context-management policies. Therefore, performance differences within each harness primarily reflect model behavior rather than differences in the evaluation setup.
% This cross-harness setting allows us to compare a code-centric harness with more general-purpose agent harnesses while keeping the underlying model fixed.
% By offering direct, containerized access to real system tools like shells and file systems, OpenClaw provides a production-grade CLI environment. Specifically, we select OpenClaw as our primary evaluation harness. To evaluate agent capabilities under complex, long-horizon tasks, we select frontier models representing the current state of the art.
% Accordingly, we evaluate six frontier models on GABench under the OpenClaw harness: Qwen3-235B, GLM-4.5-flash, GLM-5-turbo, DeepSeek-V4-pro, GPT-5-mini, and Gemini-3-flash-preview. 
% To investigate how agent performance varies between code-centric and general-purpose harnesses, 
We also perform a cross-harness evaluation using GLM-5-turbo, including 3 harnesses: OpenClaw~\cite{openclaw}, Claude Code~\cite{claudecode}, and Hermes Agent~\cite{hermes}. 
Tool schemas, system prompts, and context-management policies are kept strictly consistent within each harness, ensuring that within-harness variations in performance reflect genuine model behaviors rather than scaffold variations. Detailed experimental settings are provided in Appendix~\ref{app:experiment}.

\subsection{Performance across Tasks and Graph Types}

\input{ml_exp}
\input{Openended_exp}

\textbf{\ding{172} Graph-level retrieval tasks are more challenging than node- and edge-level retrieval tasks.}
As shown in Table~\ref{tab:retrieval_exp}, 
across all models and graph types, node-level and edge-level retrieval achieve average SRs of 34.79 and 33.05, respectively, while graph-level retrieval obtains a lower average SR of 24.54.
Compared with local node- or edge-level queries, graph-level retrieval often requires agents to extract subgraphs or aggregate statistics over larger graph regions.
These operations introduce additional requirements for parameter specification, result aggregation, and output organization, increasing the risk of errors during multi-step tool execution.

\noindent \textbf{\ding{173} Agents achieve lower performance on edge-level graph theory tasks than on node- and graph-level tasks.} Edge-level tasks obtain the lowest average SR of 23.99, compared with 35.79 for node-level and 26.77 for graph-level (Table~\ref{tab:theory_exp}).
Edge-level tasks often require agents to identify the correct node pairs, handle source--target directions when required, and interpret outputs containing paths or edge sets.
These requirements increase the risk of parameter errors, incomplete result processing, and incorrect final outputs.

\noindent \textbf{\ding{174} Agents perform better on NAGs than on TAGs in graph machine learning tasks.}
Table~\ref{tab:ml_exp} shows that node classification and link prediction achieve an average SR of 36.88 on NAGs, compared with 28.03 on TAGs.
For node classification, the average SRs on NAGs and TAGs are 23.84 and 17.66, respectively, while the corresponding link prediction SRs are 49.93 and 38.40.
A possible reason is that graph learning on TAGs requires agents to process textual attributes and align the resulting features with the graph structure before training and inference.
These additional processing and configuration steps make the workflow more error-prone.
%Table~\ref{tab:ml_exp} shows that node classification and link prediction achieve an average SR of 36.88 on NAGs, compared with 28.03 on TAGs.
%For node classification, the average SRs on NAGs and TAGs are 23.84 and 17.66, respectively, while the corresponding link prediction SRs are 49.93 and 38.40.
%A possible reason is that graph learning on TAGs requires agents to process textual attributes and align the resulting features with the graph structure before training and inference.
%These additional data-processing and configuration steps make the workflow more prone to execution errors.

\noindent \textbf{\ding{175} Graph classification has much lower tool-selection accuracy than node classification and link prediction.}
Graph classification on TPGs obtains an average TSA of 42.09, lower than 75.57 for node classification and 78.32 for link prediction. Its average SR is 24.74, which is lower than link prediction at 44.16 but slightly higher than node classification at 20.75 (Table~\ref{tab:ml_exp}). The lower scores suggest that current agents struggle to organize graph-level learning workflows and select the required tools at each step.
%For graph classification on TPGs, Table~\ref{tab:ml_exp} reports average TSA and SR scores of 42.09 and 24.74, respectively.
%Both are lower than the corresponding averages of 76.94 and 32.46 for node classification and link prediction on NAGs and TAGs.
%Since graph classification and the other two tasks are evaluated on different graph types, this comparison reflects the evaluated task--graph settings rather than the task type alone.
%Nevertheless, the lower scores suggest that current agents struggle to organize graph-level learning workflows and select the required tools at each step.

%For graph classification on TPGs, Table~\ref{tab:ml_exp} reports average TSA and SR scores of 42.09 and 24.74, respectively.
%Both are lower than the corresponding averages of 76.94 and 32.46 for node classification and link prediction on NAGs and TAGs.
%Since graph classification and the other two tasks are evaluated on different graph types, this comparison reflects the evaluated task--graph settings rather than the task type alone.
%Nevertheless, the lower scores indicate that current agents have difficulty organizing graph-level learning workflows and selecting the required tools for each step.

\begin{figure*}[t]
  \centering

  \begin{minipage}[t]{0.47\textwidth}
    \makeatletter
    \renewenvironment{table}[1][]{\def\@captype{table}}{}
    \makeatother
    \input{harness_ml}
  \end{minipage}%
  \hfill
  \begin{minipage}[t]{0.485\textwidth}
    \makeatletter
    \renewenvironment{table}[1][]{\def\@captype{table}}{}
    \makeatother
    \input{harness_openended}
  \end{minipage}

  \vskip 0.1in
  \includegraphics[width=1.0\textwidth]{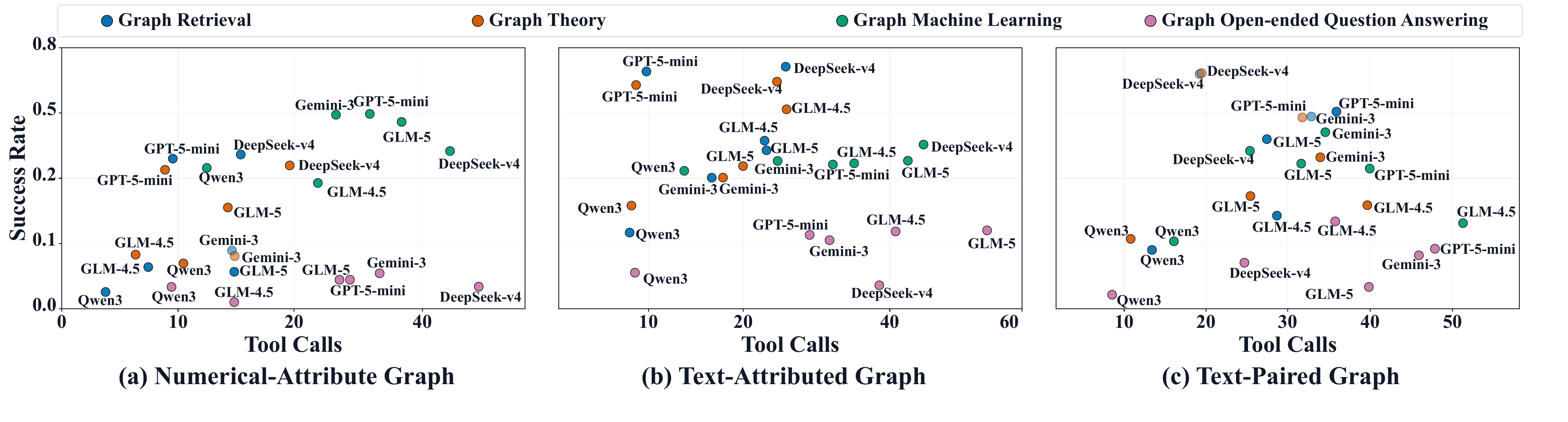}
  \vskip -0.25in
  \caption{Relationship between the average number of tool calls and SR across models and task categories.}
  \vskip -0.15in
  \label{fig:tc_sr}
\end{figure*}

\noindent \textbf{\ding{176} Agents achieve the lowest performance on graph open-ended question answering among the four task categories.}
As shown in Table~\ref{tab:openended_exp}, graph open-ended question answering obtains average SR and TSA scores of only 6.64 and 11.00, respectively, which are substantially lower than those of graph retrieval (30.79 and 49.79), graph theory (28.85 and 46.28), and graph machine learning (30.91 and 69.97).
Unlike the other three tasks, graph open-ended questions express their analysis goals through domain-specific scenarios without explicitly specifying tool names or graph operations.
The low TSA and SR suggest that agents have difficulty both inferring the required analysis operations from user requests and using the resulting evidence to produce correct final answers.

\noindent \textbf{\ding{177} DeepSeek-V4-Pro achieves the best overall performance.}
Aggregating the results reported in Tables~\ref{tab:retrieval_exp}, \ref{tab:theory_exp}, \ref{tab:ml_exp}, and~\ref{tab:openended_exp}, DeepSeek-V4-Pro achieves the highest overall SR of 37.70, followed by GPT-5-mini at 34.52. It also obtains the highest overall TSA of 61.52, slightly higher than Gemini-3-Flash at 60.87. These results show that DeepSeek-V4-Pro provides the strongest overall performance among the evaluated models under the OpenClaw harness.

\noindent \textbf{\ding{178} TAGs achieve the highest overall performance, although their advantage is task-dependent.}
A comparison of the results across Tables~\ref{tab:retrieval_exp}, \ref{tab:theory_exp}, \ref{tab:ml_exp}, and~\ref{tab:openended_exp} shows that 
TAGs obtain the highest SR of 29.43, compared with 25.42 on TPGs and 17.28 on NAGs. The advantage is mainly observed in graph retrieval, graph theory, and graph open-ended question answering, whereas NAGs perform better on graph machine learning. It suggests that node-level textual attributes may provide useful semantic information for identifying graph entities, understanding task targets, and interpreting tool outputs, but do not simplify graph learning workflows.
%A comparison of the results across Tables~\ref{tab:retrieval_exp}, \ref{tab:theory_exp}, \ref{tab:ml_exp}, and~\ref{tab:openended_exp} shows that TAGs achieve the highest SR of 29.43, compared with 25.42 on TPGs and 17.28 on NAGs.
%TAGs also obtain the highest average TSA of 48.91, while NAGs and TPGs reach 43.93 and 36.45, respectively.
%This pattern may be associated with the node-level textual attributes of TAGs, which provide direct semantic cues for identifying graph entities, understanding task targets, and interpreting tool outputs.

\noindent \textbf{\ding{179} LLM agents exhibit limited performance on challenging graph analysis tasks, especially graph machine learning and graph open-ended question answering.}
As shown in Tables~\ref{tab:ml_exp} and~\ref{tab:openended_exp}, SRs remain below 40\% in 24 of the 30 model--task--graph settings for graph machine learning and in all 36 settings for graph open-ended question answering.
The average SRs of these two task categories are only 30.91 and 6.64, respectively. 
These results indicate that current LLM agents remain unreliable on graph machine learning and graph open-ended question answering tasks.

%DeepSeek-V4-Pro obtains the highest overall SR of 37.70.
%Its average SR reaches 56.72 on graph retrieval and 52.90 on graph theory, but decreases to 33.73 on graph machine learning and 7.44 on graph open-ended question answering.
%These results indicate that strong performance on explicit retrieval and graph computation tasks does not guarantee reliable execution of graph learning workflows or scenario-based open-ended graph analysis.

\subsection{Impact of Agent Harnesses}
% Agent harnesses differ in their support for tool invocation and iterative execution, which may cause the same backbone LLM to exhibit different graph analysis performance across harnesses. We therefore conduct a cross-harness comparison on two challenging task categories---graph machine learning and graph open-ended question answering. We use GLM-5-Turbo as the backbone LLM since it is supported by all three harnesses.
Agent harnesses provide different levels of support for tool invocation and iterative execution. We therefore compare three harnesses on two challenging task categories: graph machine learning and graph open-ended question answering. We use GLM-5-turbo as the backbone LLM because it is supported by all harnesses.
%Graph machine learning requires end-to-end training and inference, while graph open-ended question answering shows the lowest overall performance. We therefore compare agent harnesses on these two challenging categories using GLM-5-Turbo, which is supported by all three harnesses.

%%79
%DeepSeek-V4-Pro ranks first on graph retrieval and graph theory, with average SRs of 56.72 and 52.90, respectively. However, its SR decreases to 33.73 on graph machine learning and 7.44 on graph open-ended question answering. These results indicate that strong performance on explicit retrieval and graph computation tasks does not guarantee reliable execution of graph machine learning workflows or scenario-based open-ended graph analysis.

\noindent \textbf{\ding{180} Claude Code performs more reliably than Hermes and OpenClaw on both task categories.}
As shown in Tables~\ref{tab:harness_ml} and~\ref{tab:harness_openended}, Claude Code achieves an average SR of 87.80 on graph machine learning, compared with 60.22 for Hermes and 34.94 for OpenClaw. It also reaches 16.71 on graph open-ended question answering, outperforming OpenClaw at 10.28 and Hermes at 8.17. This advantage may arise from its support for iterative code and tool execution, which aligns well with the multi-step nature of these tasks.

\subsection{Efficiency Analysis}
\label{sec:efficiency}
To examine agent execution efficiency, we analyze the relationship between task success rate and the average number of tool calls. All models are evaluated using the OpenClaw harness with the same toolset and task environment. Additional analyses of token consumption and token efficiency across models and task categories are provided in Appendix~\ref{app:token_analysis}.

\noindent \textbf{\ding{181} Longer tool-call chains do not necessarily improve task success.}
Figure~\ref{fig:tc_sr} shows that on TAG retrieval, GPT-5-mini achieves an average SR of 69.09 with about 10 tool calls, while GLM-5-turbo and GLM-4.5-flash use around 22 calls but reach only 32.88 and 37.29, respectively.
Similarly, on TPG retrieval, DeepSeek-V4-Pro achieves an SR of 67.95 with about 20 calls, outperforming several models using around 30 calls.
These results suggest that success depends more on effective tool selection, parameter configuration, and use of intermediate results than on the number of calls.

\subsection{Summary and Future Work}
Based on the above experimental results and observations, we summarize three key conclusions across the dimensions of task capability, agent harness, and tool use:
(1) Existing LLM agents still struggle with complex graph analysis tasks, particularly graph machine learning and graph open-ended question answering, where success rates are below 40\% in most cases. These results highlight their limited ability to understand structures and reason over graph data. Future research should develop more graph-aware agents. One promising direction is to integrate stronger graph models, such as graph foundation models~\cite{liu2025graph}, with the task-planning and tool-use capabilities of LLM agents.
(2) Harnesses significantly affect the graph analysis performance of LLM agents, but existing harnesses remain limited on complex graph tasks. 
% Harness choice significantly affects performance, yet existing harnesses remain limited on complex graph tasks.
% Harness choice strongly affects performance, yet existing harnesses struggle with complex graph analysis.
% For the same GLM-5-Turbo, the average success rate on graph machine learning tasks differs by more than 50 percentage points across harnesses. Nevertheless, even ClaudeCode, the best-performing harness, achieves a success rate of only 29.63\% on open-ended graph question answering. 
This suggests that general harnesses can facilitate well-structured multi-step execution but remain inadequate for complex graph analysis tasks, highlighting the need to develop graph-specific agent harnesses.
(3) Tool-call quality is more important than tool-call quantity for successful graph analysis. 
% Graph analysis depends more on tool-call quality than quantity.
% (3) Tool-call quality is more important for successful graph analysis than tool-call quantity. 
% Some models achieve substantially higher success rates with fewer tool calls than models with longer execution trajectories, indicating that additional calls cannot compensate for poor execution decisions. 
Therefore, future agent research for graph analysis should prioritize graph-tool routing, parameter validation, and trajectory-level execution strategies rather than simply expanding the toolset or increasing the tool-call budget.

% existing harnesses still struggle to support complex graph analysis tasks.

%  across the dimensions of task capability, Agent Harness, and tool use.

% Overall, our evaluation leads to three main observations:

% (1) Challenging graph analysis remains difficult for existing LLM agents regardless of the agent harness, particularly for graph open-ended question answering, where all success rates are lower than 30\%.
% (2) Agent performance depends strongly on the surrounding harness, as the same backbone model shows a performance gap exceeding 60 percentage points across harnesses on graph machine learning tasks.
% (3) Longer tool-use trajectories are not inherently more effective; capable models can complete tasks more successfully while invoking fewer tools.

%% file: ml_exp.tex
\begin{table*}[t]
\centering
\vskip -0.05in
\caption{LLM-agent performance on graph machine learning tasks. Best and second-best results are \textbf{bolded} and \underline{underlined}.}
\label{tab:ml_exp}
\vskip -0.15in
\resizebox{\textwidth}{!}{%
\setlength{\tabcolsep}{7pt}%
\renewcommand{\arraystretch}{0.4}%
\newcommand{\mlbodystrut}{\rule[-4pt]{0pt}{13pt}}%
\begin{tabular}{l *{10}{>{\centering\arraybackslash}m{0.82cm}}}
\toprule
\multirow{3}{*}[-6pt]{\textbf{Model}}
& \multicolumn{4}{c}{\textbf{Numerical-Attribute Graph}}
& \multicolumn{4}{c}{\textbf{Text-Attributed Graph}}
& \multicolumn{2}{c}{\textbf{Text-Paired Graph}} \\
\cmidrule(lr){2-5}
\cmidrule(lr){6-9}
\cmidrule(lr){10-11}
& \multicolumn{2}{c}{Node Classification}
& \multicolumn{2}{c}{Link Prediction}
& \multicolumn{2}{c}{Node Classification}
& \multicolumn{2}{c}{Link Prediction}
& \multicolumn{2}{c}{Graph Classification} \\
\cmidrule(lr){2-3}
\cmidrule(lr){4-5}
\cmidrule(lr){6-7}
\cmidrule(lr){8-9}
\cmidrule(lr){10-11}
& SR & TSA & SR & TSA & SR & TSA & SR & TSA & SR & TSA \\
\midrule
\mlbodystrut Qwen3-235B
& \underline{27.70} & 25.76
& 21.82 & 57.35
& \textbf{22.23} & 30.18
& 24.63 & 26.87
& 10.34 & 14.48 \\

\mlbodystrut GLM-4.5-flash
& 8.57 & 67.66
& 30.00 & 71.23
& 15.68 & 77.94
& \underline{38.09} & 85.27
& 13.13 & 35.50 \\

\mlbodystrut GLM-5-turbo
& 26.66 & \textbf{99.99}
& 65.18 & \textbf{99.99}
& \underline{19.36} & 83.63
& 36.77 & 87.61
& 26.74 & \textbf{74.39} \\

\mlbodystrut DeepSeek-V4-pro
& 20.88 & \underline{96.01}
& 44.18 & \underline{99.03}
& 14.54 & \underline{90.02}
& \textbf{56.45} & \underline{90.96}
& \underline{32.61} & 40.25 \\

\mlbodystrut GPT-5-mini
& \textbf{32.54} & 79.75
& \underline{66.66} & 68.11
& 15.43 & \textbf{99.60}
& 37.23 & \textbf{97.79}
& 24.46 & 42.78 \\

\mlbodystrut Gemini-3-flash
& 26.66 & 85.68
& \textbf{71.73} & 88.32
& 18.71 & 70.65
& 37.22 & 67.27
& \textbf{41.18} & \underline{45.11} \\
\bottomrule
\end{tabular}%
}
\vskip -0.05in
\end{table*}

%% file: Openended_exp.tex
\begin{table*}[t]
\centering
\caption{Performance of LLM agents on graph open-ended question-answering tasks. The best and second-best results in each column are shown in \textbf{bold} and \underline{underlined}, respectively.}
\label{tab:openended_exp}
\vskip -0.15in
\resizebox{\textwidth}{!}{%
\setlength{\tabcolsep}{4pt}%
\begin{tabular}{l *{12}{>{\centering\arraybackslash}m{0.82cm}}}
\toprule
\multirow{3}{*}[-2pt]{\textbf{Model}}
& \multicolumn{4}{c}{\textbf{Numerical-Attribute Graph}}
& \multicolumn{4}{c}{\textbf{Text-Attributed Graph}}
& \multicolumn{4}{c}{\textbf{Text-Paired Graph}} \\
\cmidrule(lr){2-5}
\cmidrule(lr){6-9}
\cmidrule(lr){10-13}
\noalign{\vskip -1pt}
& \multicolumn{2}{c}{Insight}
& \multicolumn{2}{c}{Prediction}
& \multicolumn{2}{c}{Insight}
& \multicolumn{2}{c}{Prediction}
& \multicolumn{2}{c}{Insight}
& \multicolumn{2}{c}{Prediction} \\[-1pt]
\cmidrule(lr){2-3}
\cmidrule(lr){4-5}
\cmidrule(lr){6-7}
\cmidrule(lr){8-9}
\cmidrule(lr){10-11}
\cmidrule(lr){12-13}
\noalign{\vskip -1pt}
& SR & TSA & SR & TSA
& SR & TSA & SR & TSA
& SR & TSA & SR & TSA \\[-4pt]
\midrule
Qwen3-235B
& 2.35
& \textbf{24.05}
& \textbf{4.36}
& 2.69
& 5.33
& \textbf{27.53}
& 5.74
& 1.90
& 0.60
& 1.31
& 3.70
& 1.77 \\

GLM-4.5-flash
& 0.34
& 1.39
& 0.30
& 2.03
& 3.92
& 6.84
& 3.28
& 0.69
& 0.57
& 4.05
& 6.12
& 2.75 \\

GLM-5-turbo
& \underline{7.65}
& 6.72
& \underline{3.22}
& \textbf{22.14}
& \underline{13.56}
& 9.90
& \textbf{10.47}
& \textbf{24.86}
& \textbf{14.29}
& 9.53
& \textbf{12.50}
& \underline{16.65} \\

DeepSeek-V4-pro
& 4.80
& \underline{14.22}
& 1.98
& \underline{17.44}
& \textbf{14.91}
& \underline{10.02}
& 8.80
& \underline{18.54}
& 10.81
& \textbf{14.84}
& 3.33
& 14.82 \\

GPT-5-mini
& 5.92
& 2.80
& 2.96
& 8.16
& 12.91
& 6.76
& \underline{9.72}
& 10.76
& 10.11
& \underline{10.38}
& \underline{8.29}
& 5.73 \\

Gemini-3-flash
& \textbf{7.76}
& 2.73
& 1.15
& 12.19
& 12.94
& 3.46
& 8.08
& 12.80
& \underline{11.63}
& 6.98
& 4.76
& \textbf{56.66} \\
\bottomrule
\end{tabular}%
}
\vskip -0.1in
\end{table*}

%% file: harness_ml.tex
% Required packages:
% \usepackage{booktabs}
% \usepackage{multirow}
% \usepackage{array}

\begin{table}[t]
\vskip -0.05in
\centering
\caption{Task success rate (SR, \%) of GLM-5-Turbo under different agent harnesses on graph machine learning tasks. The best and second-best results in each column are shown in \textbf{bold} and \underline{underlined}, respectively.}
\label{tab:harness_ml}
\vskip 0.02in

\normalsize
\setlength{\tabcolsep}{2.0pt}
\renewcommand{\arraystretch}{1.12}

\begin{tabular*}{\columnwidth}{
    @{\extracolsep{\fill}}
    >{\centering\arraybackslash}m{1.65cm}
    c c c c c
    @{}
}
\toprule

\multirow[c]{2}{*}{\textbf{Harness}}
& \multicolumn{2}{c}{\textbf{NAGs}}
& \multicolumn{2}{c}{\textbf{TAGs}}
& \textbf{TPGs} \\[-2pt]

\cmidrule(lr){2-3}
\cmidrule(lr){4-5}
\cmidrule(l){6-6}

& \textbf{NC}
        & \textbf{LP}
        & \textbf{NC}
        & \textbf{LP}
        & \textbf{GC} \\[-2pt]
\midrule

Hermes
& \underline{61.54}
& \textbf{90.90}
& \underline{54.35}
& \underline{87.18}
& 7.14 \\

Claude Code
& \textbf{95.23}
& \underline{78.57}
& \textbf{86.79}
& \textbf{89.53}
& \textbf{88.89} \\

OpenClaw
& 26.66
& 65.18
& 19.36
& 36.77
& \underline{26.74} \\

\bottomrule
\end{tabular*}

\vskip -0.06in
\end{table}

%% file: harness_openended.tex
% Required packages:
% \usepackage{booktabs}
% \usepackage{multirow}
% \usepackage{array}
% \usepackage{graphicx}

\begin{table}[t]
\centering
\vskip -0.05in
\caption{Task success rate (SR, \%) of GLM-5-Turbo under different agent harnesses on graph open-ended question-answering tasks. The best and second-best results in each column are shown in \textbf{bold} and \underline{underlined}, respectively.}
\label{tab:harness_openended}
\vskip 0.02in

\normalsize
\setlength{\tabcolsep}{1pt}
\renewcommand{\arraystretch}{1.12}

\begin{tabular*}{\columnwidth}{
    @{\extracolsep{\fill}}
    >{\centering\arraybackslash}m{1.65cm}
    *{6}{c}
    @{}
}
\toprule

\multirow[c]{2}{*}{\textbf{Harness}}
& \multicolumn{2}{c}{\textbf{NAGs}}
& \multicolumn{2}{c}{\textbf{TAGs}}
& \multicolumn{2}{c}{\textbf{TPGs}} \\[-2pt]

\cmidrule(lr){2-3}
\cmidrule(lr){4-5}
\cmidrule(lr){6-7}

& {\small\textbf{Insight}}
& {\small\textbf{Prediction}}
& {\small\textbf{Insight}}
& {\small\textbf{Prediction}}
& {\small\textbf{Insight}}
& {\small\textbf{Prediction}} \\[-2pt]

\midrule

Hermes
& 0.00
& \textbf{8.22}
& 1.28
& \underline{15.12}
& \underline{16.28}
& 8.11 \\

Claude Code
& \underline{7.63}
& \underline{3.70}
& \textbf{18.59}
& \textbf{20.00}
& \textbf{29.63}
& \textbf{20.69} \\

OpenClaw
& \textbf{7.65}
& 3.22
& \underline{13.56}
& 10.47
& 14.29
& \underline{12.50} \\

\bottomrule
\end{tabular*}

\vskip -0.08in
\end{table}

%% file: conclusion.tex
%In this paper, we introduced GABench, a large-scale benchmark for evaluating LLM agents on real-world graph analysis tasks. GABench covers four task categories across three graph types and includes 10,400 graph analysis questions with verifiable ground truth, together with an executable toolset and graph analysis environment. Our evaluation of trajectory-level tool use and task completion shows that current LLM agents still struggle with challenging graph analysis tasks, indicating that further improvements are needed in their graph analysis capabilities.
% In this paper, we introduced GABench, a large-scale benchmark for evaluating LLM agents on real-world graph analysis tasks. It contains 10,400 verifiable questions across four task categories and three graph types, and provides an executable toolset and graph environment. Our evaluation shows that current LLM agents still struggle with challenging graph analysis tasks.
% \section{Conclusion}
In this paper, we introduce GABench, a comprehensive benchmark for evaluating LLM agents on graph analysis tasks.
GABench integrates 13 real-world datasets spanning 6 domains and 3 graph types, covers 4 graph analysis task categories, and provides 84 executable tools.
Based on these resources, we construct 10,400 executable tasks with verifiable ground truth to evaluate agents' planning, tool use, and end-to-end task completion in graph environments.
Extensive evaluations of frontier LLMs and agent harnesses show that current agents still struggle with complex graph analysis, particularly graph machine learning and graph open-ended question answering.
We further find that harness choice substantially affects performance and that successful graph analysis depends more on tool-call quality than quantity.
Overall, GABench provides a unified testbed and practical guidance for developing more capable LLM agents and graph-specific agent harnesses.

%% file: related_work.tex
\subsection{Evaluating LLMs with Graph Tasks}
Graph tasks require LLMs to process graph structures, node semantics, and multi-step dependencies, making them an important setting for model evaluation~\cite{DBLP:journals/corr/abs-2602-06319,xu2025graphomni,luo2024graphinstruct,tang2024grapharena,yuan2024gracore,chen2024graphwiz}. Early benchmarks, such as NLGraph~\cite{wang2023can}, GPT4Graph~\cite{guo2023gpt4graph}, and GraphWiz~\cite{chen2024graphwiz}, mainly use synthetic graphs to evaluate basic graph understanding and reasoning. GraphInstruct~\cite{luo2024graphinstruct} and GraphOmni~\cite{xu2025graphomni} extend the evaluation to a broader range of graph-theoretic tasks at the node, edge, and graph levels. More recent benchmarks further improve the realism and scope of evaluation. ProGraph~\cite{DBLP:conf/nips/LiCCLSLQW000Y24} introduces graph-theoretic problems grounded in real-world graphs, while GrAlgoBench~\cite{DBLP:journals/corr/abs-2602-06319} formulates graph retrieval and graph computation as open-ended questions in real-world contexts.
Despite this progress, existing benchmarks primarily evaluate LLMs in static question-answering settings and do not assess whether agents can plan, select and use tools, and interact with external graph environments. Moreover, existing benchmarks cover a limited range of graph types and tasks.
% Moreover, no existing benchmark jointly covers all four major task families considered in our work: graph retrieval, graph theory, graph machine learning, and open-ended graph question answering. 
GABench addresses these limitations by providing a unified benchmark for evaluating LLM agents across diverse real-world graph analysis tasks.

% Real-world graph problems require large language models to possess a deep understanding of graph structures and node semantics, alongside multi-hop reasoning capabilities, making them a significant challenge for model evaluation. Existing benchmarks utilize synthetic graph data to evaluate models, as demonstrated by frameworks like NLGraph, GPT4Graph, and GraphWiz. Other benchmarks, including GraphInstruct and GraphOmni, provide a broader evaluation of graph theoretic problems from the node level to the graph level. Furthermore, ProGraph incorporates real-world graph theoretic problems into evaluation frameworks, while GRALGOBENCH frames these problems and graph retrieval tasks as real-world graph open-ended question answering to evaluate large language models. Despite these efforts, no existing work specifically evaluates large language model agents. Current benchmarks only cover a single category among graph retrieval, graph theory, graph machine learning, and open-ended question answering. This limitation motivates us to develop a unified benchmark that encompasses all four categories of real-world graph tasks to evaluate agents and provide comprehensive insights.

\subsection{Benchmarking LLM Agents}
LLM agents extend beyond standard question answering by planning and executing multi-step actions in external environments~\cite{Liu-ICLR2024,Xie-ICML2024,Patil-ICML2025,Deng-NeurIPS2023,Yan-arXiv2025,Liu-arXiv2025,ding2026wildclawbenchbenchmarkrealworldlonghorizon}. Existing agent benchmarks cover a wide range of settings, including software engineering with SWE-bench~\cite{jimenez2023swe}, Terminal-Bench~\cite{merrill2026terminalbenchbenchmarkingagentshard}, and LiveCodeBench~\cite{zheng2025livecodebench}; web and GUI interaction with WebArena~\cite{Zhou-ICLR2024}, WebShop~\cite{yao2022webshop}, and VisualWebArena~\cite{Koh-arXiv2024}; web research with BrowseComp~\cite{wei2025browsecomp}; and general tool use with MCP-Bench~\cite{wang2025mcpbenchbenchmarkingtoolusingllm} and $\tau$-bench~\cite{Yao-ICLR2025}. However, these benchmarks rarely consider graph analysis tasks or graph-based environments. As a result, they cannot fully assess how agents plan and use tools to solve practical graph problems. This gap motivates GABench, which uses real-world graph analysis tasks to evaluate the planning and tool-use capabilities of LLM agents.

% Large language models increasingly drive autonomous agents, which expand beyond standard question answering to execute multi-step operations across diverse environments on behalf of users. Agent benchmarks are primarily categorized by their interaction interfaces, spanning domains such as software engineering with SWE-bench, Terminal-Bench, and LiveCodeBench; web and GUI control with WebArena, WebShop, and VisualWebArena; browser-centric research with BrowseComp; and tool orchestration with MCP-Bench and tau-bench. Although tool orchestration benchmarks focus on real-world tasks, they neglect graph data as a crucial element of practical scenarios. This omission inspires us to utilize real-world graph problems to evaluate the tool orchestration and planning capabilities of agents.

%% file: Appendix.tex
% Keep this line only if \appendix is not called in main.tex.
\section{Datasets Overview}
\label{app:datasets}

To evaluate LLM agents across diverse graph structures and real-world application settings, GABench includes datasets that vary substantially in graph scale, attribute type, and domain. Table~\ref{tab:graph_statistics} summarizes the statistics and domains of the 13 real-world datasets used in GABench. The six text-attributed graph datasets each contain a single graph, ranging from 2,708 to 2,449,029 nodes and from 10,556 to 123,618,192 edges, and cover social networks, e-commerce networks, and academic citation networks. The five numerical-attribute graph datasets also each contain a single graph, ranging from 403 to 3,700,550 nodes and from 16,570 to 4,300,999 edges, and span the finance, social, and justice domains. The two text-paired graph datasets consist of molecular graph collections: HIV contains 41,127 graphs and PCBA contains 437,092 graphs, with each graph containing approximately 26 nodes and 55--56 edges. Overall, these datasets cover six real-world domains and provide graph structures with substantially different numbers of graphs, nodes, and edges.

Specifically, the TAG datasets comprise Reddit~\cite{li2024glbenchcomprehensivebenchmarkgraph}, Instagram~\cite{li2024glbenchcomprehensivebenchmarkgraph}, History~\cite{chen2024textspacegraphfoundationmodels}, Product~\cite{chen2024llaga}, Arxiv~\cite{chen2024llaga}, and Cora~\cite{chen2024llaga}. The NAG datasets comprise DGraph-Fin~\cite{huang2023dgraphlargescalefinancialdataset}, Pokec~\cite{dai2021saydiscriminationlearningfair}, NBA~\cite{dai2021saydiscriminationlearningfair}, Credit~\cite{Ma_2022}, and Bail~\cite{Ma_2022}, while the TPG datasets comprise HIV~\cite{chen2024textspacegraphfoundationmodels} and PCBA~\cite{chen2024textspacegraphfoundationmodels}.
\input{graph_statistics}
\section{Experiment Details}
\subsection{Details of Evaluated Models}
\label{app:experiment}
In this section, we report the roles and decoding configurations of all LLMs used in GABench, including the evaluated agents, the question generator, and the judge model. Table~\ref{tab:model_temp} lists the temperature and top-$p$ settings for each model.
\input{model_temp} 
\subsection{LLM-as-a-Judge Evaluation Protocol}
\label{app:judge_protocol}
\input{token_retrieval}
\input{token_theory}
We employ DeepSeek-V4-Flash as the judge model. For each evaluation instance, the judge receives the original question, the ground-truth answer, and the final response generated by the agent. An LLM-based evaluator is used because graph open-ended question-answering responses may combine multiple findings and explanations in diverse formats, making semantically equivalent answers difficult to assess using exact-match rules. Moreover, even in graph retrieval, graph theory, and graph machine learning tasks with verifiable answers, agent responses may contain intermediate hypotheses, tentative conclusions, and self-corrections, causing rule-based methods to mistake intermediate results for the final answer. The judge therefore identifies the agent's final conclusion and assesses its semantic consistency with the ground truth, while ignoring differences in wording, format, or presentation order that do not affect correctness. A score of $1$ indicates successful task completion, while a score of $0$ indicates failure. Infrastructure failures, such as API or connection errors, are separately assigned a score of $-1$. Instances assigned a score of $-1$ are either reviewed by human annotators or rerun to obtain a valid agent response.

\section{Token Efficiency Analysis}
\label{app:token_analysis}
\input{token_ml}
\input{token_openend}
\subsection{Token Measurement Protocol}
\label{app:token_measurement}
To provide a detailed view of LLM-agent efficiency, we separately measure the input and output tokens consumed by the agent for each task. For task
$i$, let $m_i$ denote the number of LLM calls made by the agent,
and let $T_{\mathrm{in}}^{(i,j)}$ and
$T_{\mathrm{out}}^{(i,j)}$ denote the input and output tokens of
the $j$-th call, respectively. 
%The average per-task input and output token consumption over $N$ tasks are computed as
The average numbers of input and output tokens consumed per task over $N$ tasks are computed as follows:
\begin{equation}
\bar{T}_{\mathrm{in}}
=
\frac{1}{N}
\sum_{i=1}^{N}
\sum_{j=1}^{m_i}
T_{\mathrm{in}}^{(i,j)},
\qquad
\bar{T}_{\mathrm{out}}
=
\frac{1}{N}
\sum_{i=1}^{N}
\sum_{j=1}^{m_i}
T_{\mathrm{out}}^{(i,j)}.
\end{equation}
The reported token consumption therefore reflects the complete
agent trajectory rather than only the final response. Input and
output tokens are reported separately because input consumption is
also affected by context accumulation, tool schemas, intermediate
observations, and the context-management strategy of the harness.

\subsection{Token Consumption across Tasks and Models}
\label{app:token_across_tasks}

\noindent \textbf{\ding{172} Token consumption varies substantially across
models under the same agent harness.}
The differences are particularly pronounced in input-token
consumption. On graph retrieval and graph theory, GLM-5-Turbo
uses an average of 24,528 and 19,455 input tokens, respectively,
whereas Gemini-3-Flash uses 292,541 and 258,992. The same pattern
appears in graph machine learning, where Gemini-3-Flash consumes
492,259 input tokens on average, compared with 61,434 for
GLM-5-Turbo and 56,786 for DeepSeek-V4-Pro. These differences
suggest that models follow substantially different interaction
patterns even when they use the same harness, toolset, and task
environment. 

\noindent\textbf{\ding{173} Graph machine learning and open-ended
question answering incur higher token costs than graph retrieval
and graph theory.}
Averaged across the six models, graph machine learning and
open-ended question answering consume approximately 192,693 and
186,014 input tokens per task setting, respectively, compared with
109,746 for graph retrieval and 100,117 for graph theory. Graph
open-ended question answering also produces the largest output,
averaging 12,737 output tokens, whereas graph retrieval, graph
theory, and graph machine learning average 5,270, 5,743, and
7,021, respectively. This pattern is consistent with the longer
workflows required by graph machine learning and the additional
interpretation and response synthesis required by open-ended
questions.
\input{tokenml_harness}

\input{tokenopen_harness}
\begin{figure*}[t]
    \centering
    \includegraphics[width=\textwidth]{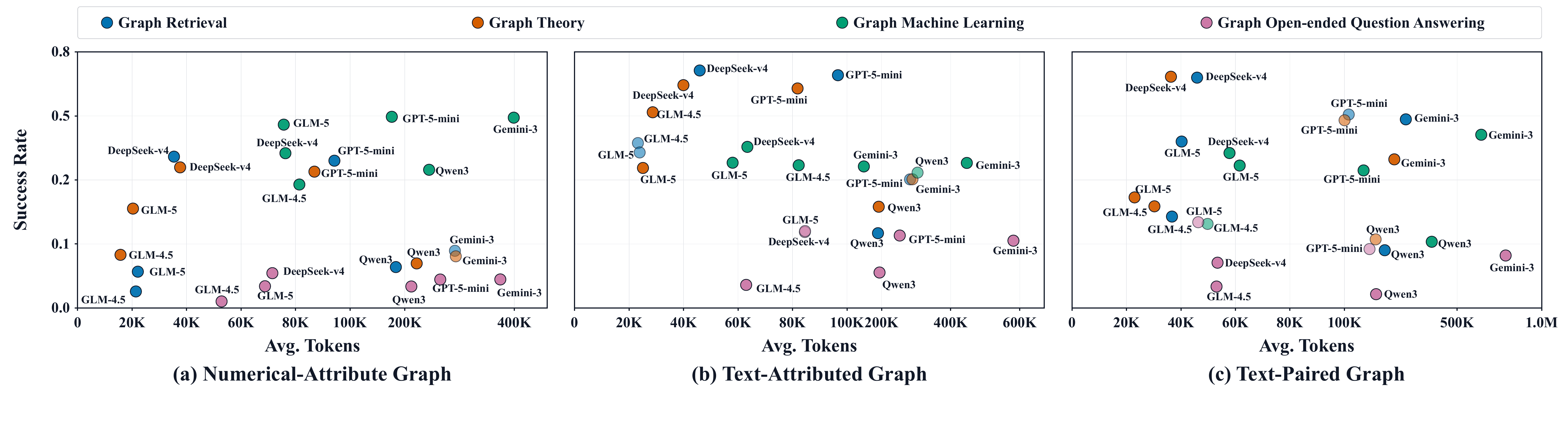}
    \caption{Relationship between average input-token consumption
    and task success rate (SR) across models and task categories.
    Panels (a)--(c) report results on numerical-attribute,
    text-attributed, and text-paired graphs, respectively.
    Each point represents one model--task-category pair, and
    marker colors distinguish the four task categories.}
    \label{fig:token_sr}
\end{figure*}
\noindent\textbf{\ding{174} Prediction questions are more
token-intensive than insight questions in open-ended graph
analysis.}
Averaged across models and graph types, prediction tasks consume
approximately 204,462 input tokens, compared with 167,565 for
insight tasks, representing an increase of 22.0\%. Their average
output consumption also increases from 12,081 to 13,393 tokens.
This result indicates that prediction questions generally require
agents to retain more intermediate context and generate more
extensive responses than questions centered on graph insights.

\noindent\textbf{\ding{175} Higher token consumption does not consistently improve task success.}
As shown in Figure~\ref{fig:token_sr}, models that consume more tokens do not systematically achieve higher SR across graph types and task categories. Gemini-3-flash often consumes among the most tokens but obtains only moderate or low SR, whereas DeepSeek-V4-Pro and GPT-5-mini frequently achieve higher SR with fewer tokens. This pattern is particularly evident in graph open-ended question answering, where SR remains below approximately 12\% despite large differences in token consumption. These results suggest that effective planning, tool selection, and use of intermediate results matter more than longer interactions.

\subsection{Impact of Agent Harness}
\noindent \textbf{\ding{176} Agent harnesses differ substantially in input-token efficiency.}
Claude Code consumes the fewest input tokens on both graph machine learning and graph open-ended question answering, averaging 15,793 and 22,849 tokens, respectively. These values are 47.0\% and 35.4\% lower than Hermes, and 74.3\% and 59.6\% lower than OpenClaw, suggesting that harness design strongly affects context and intermediate-state overhead.

\noindent \textbf{\ding{177} Graph open-ended question answering requires substantially more output tokens than graph machine learning.}
Across all three harnesses, average output consumption increases from 4,044--4,826 tokens on graph machine learning to 8,002--12,167 tokens on graph open-ended question answering. The increase is largest for OpenClaw, whose output consumption rises from 4,201 to 12,167 tokens, indicating the greater generation cost of synthesizing scenario-based graph findings.
\begin{figure*}[h]
\begin{tcolorbox}[top=2pt, bottom=2pt, left=4pt, right=4pt, halign title=center, title={Real-world Scenario Generation Prompt}]
\textbf{System prompt:} You are an expert in designing high-level real-world contexts for graph datasets. Given a graph domain context, your task is to generate abstract, domain-wide background scenarios describing general situational environments. These scenarios must describe overall domain backgrounds without being phrased as specific analytical tasks, investigative questions, or concrete queries. \\
\textbf{Rules:} Domain-Level: Focus on broad domain background without including specific node/edge entities or names. No Task/Query: Describe situational background states rather than actionable tasks (avoid verbs like "analyze", "find", or "investigate"). Output Format: Return strictly a JSON array of strings: ["Scenario 1", "Scenario 2", ...].\\
\textbf{User content:} General graph context: \{graph\_context\}. Please generate 5 to 10 abstract, high-level real-world background scenarios for this domain. 
\end{tcolorbox}
\end{figure*}

\begin{figure*}[h]
\begin{tcolorbox}[top=2pt, bottom=2pt, left=4pt, right=4pt, halign title=center, title={Fuzzy Tool Description Generation Prompt}]
\textbf{System prompt:} You are an expert in graph analytics and natural language generation. Your task is to rewrite generic graph-theory tool descriptions into domain-specific, natural-language "fuzzy" descriptions that feel native to the provided graph context.
Rules:
1. For tools requiring specific node inputs, you MUST include the exact placeholders \{node1\_placeholder\} and \{node2\_placeholder\} in the description.
2. For global tools, keep descriptions broad and domain-level without using specific node/edge sample names.
3. Output only a valid JSON object mapping each tool name to its fuzzy description: \{"tool\_name": "fuzzy description", ...\}.
4. Keep each description concise (1-3 sentences).
5. CRITICAL: Do NOT use any algorithm names or graph-theory terminology (e.g., "k-core", "shortest path", "graph", "node", "edge"). Use domain-appropriate synonyms such as "network", "system", "entities", "connections", or "relationships". \\
\textbf{User content:}
Graph Overview:
\{TAG\_subgraph\_description\}
\{graph\_level\_Summary\} \\
\textbf{Tools to Rewrite:}
\{tool\_list\}. 
Please generate domain-adaptive, fuzzy descriptions for the listed graph tools and output a single JSON object.
\end{tcolorbox}
\end{figure*}

\begin{figure*}[h]
\begin{tcolorbox}[top=2pt, bottom=2pt, left=4pt, right=4pt, halign title=center, title={Graph Analysis Question Generation Prompt}]
\textbf{System prompt:} You are an expert AI benchmark architect. Your task is to write ONE natural-language analytical question that a domain practitioner would ask, based on a pre-defined sequence of graph analytical steps. Output ONLY the final question text without any preamble or explanation. \\
\textbf{Rules:} Cascading Storytelling: Weave the analytical steps into a seamless causal narrative where each step naturally sets up the next. No Sequence Words: DO NOT use sequencing words (e.g., "first", "then", "next", "finally"). Domain Language: Describe the domain intent of each step. Do NOT mention any internal tool or algorithm names. Format: Require the final results in JSON format. \\
\textbf{User content:} Background: Real-World Scenarios in this domain: "\{real\_world\_scenarios\}". Tool Descriptions: "\{tool\_descriptions\}" 
\end{tcolorbox}
\end{figure*}

\subsection{Summary and Future Work}
Overall, our token analysis shows that efficiency is affected by both the backbone model and the agent harness. Even under the same harness and task environment, models exhibit large differences in input-token consumption, while different harnesses introduce substantially different context-management overhead. Graph machine learning and graph open-ended question answering are generally more token-intensive than graph retrieval and graph theory, with prediction questions requiring more tokens than insight questions. Among the evaluated harnesses, Claude Code achieves the lowest average input-token consumption, whereas OpenClaw incurs the highest cost in most settings. Future work should therefore investigate more token-efficient agent designs, including compact intermediate-state representations, selective retention of tool observations, adaptive context pruning, and early termination of redundant tool calls.

\section{Details of the Prompts Used}

\subsection{Graph Analysis Question Generation Prompt}
\label{app:generation_prompt}
In this section, we show the detailed prompt used for graph analysis question generation in GABench.

\subsection{LLM-as-a-Judge Prompt}
\label{app:judge_prompt}
% Add these packages to the preamble:
In this section, we show the detailed prompt used for LLM-as-a-Judge in GABench.
% Optional box color
\begin{figure*}[t]
\begin{tcolorbox}[top=2pt, bottom=2pt, left=4pt, right=4pt, halign title=center, title={LLM-as-a-Judge Prompt}]
You are evaluating the final output of an AI agent solving a graph/dataset task.\\
\textbf{User Question:} \{user\_question\}\\
\textbf{Task Category:} \{args.task\}\\
\textbf{Ground Truth:} \{args.ground\_truth\}\\
\textbf{Note:} Large lists (like `subgraph\_nodes' or `filtered\_nodes') in the ground truth are masked with `[\texttt{MASKED\_LIST}...]' to save space.\\
For masked fields, just verify whether the agent’s output shows evidence of correctly completing that step/behavior (e.g., if the agent's output has the key and a list). Do not penalize for exact element mismatches if the elements are masked.\\
\{json.dumps\}\\
\textbf{Agent's Final Output:}\\
\{agent\_output\_to\_eval\}\\
Evaluate the agent's output against the Ground Truth.\\
Assign a score of  1 (Pass), 0 (Fail), or -1 (infrastructure failure). If the agent's output correctly matches our ground truth (Pass), assign 1. If it does not match (Fail), assign 0. If the agent's output indicates an API request error, connection error, or similar failure, assign -1.\\
Respond ONLY with a valid JSON object in this format:\\
\{\\
\hspace*{1em}"score": \textless{}int, 1, 0, or -1\textgreater{},\\
\hspace*{1em}"reason": \textless{}string explaining the evaluation details\textgreater{}\\
\}
\end{tcolorbox}
\end{figure*}

\section{Case Study}
This section presents representative cases from the Cora dataset for the four task categories in GABench: graph retrieval, graph theory, graph machine learning, and graph open-ended question answering. For each category, we report the user question and the corresponding agent output to illustrate how LLM agents solve different graph analysis tasks.

\begin{figure*}[t]
\begin{tcolorbox}[
    top=2pt,
    bottom=2pt,
    left=4pt,
    right=4pt,
    halign title=center,
    title={Graph Retrieval Case Study}
]
\textbf{User Question}: Extract the 3-hop ego subgraph centered at node 28. How many nodes are in this subgraph?

\medskip
\textbf{Agent Output}: The 3-hop ego subgraph centered at node 28 in the Cora dataset contains \textbf{180 nodes} and 285 edges. The subgraph was extracted successfully without reaching the default 500-node \texttt{max\_nodes} limit, so no truncation was required.
\end{tcolorbox}
\end{figure*}

\begin{figure*}[t]
\begin{tcolorbox}[
    top=2pt,
    bottom=2pt,
    left=4pt,
    right=4pt,
    halign title=center,
    title={Graph Theory Case Study}
]
\textbf{User Question}: Extract the 3-hop ego subgraph centered at node 434. What is the dominating set of this subgraph?

\medskip
\textbf{Agent Output}: Now let me sort these nodes in ascending order and format the answer. The dominating set of the 3-hop ego subgraph centered at node 434, with \texttt{max\_nodes=500}, is:

\begin{quote}
\small\ttfamily
[106, 202, 258, 274, 321, 434, 482, 523, 634, 741, 875, 885, 944, 1024, 1032, 1101, 1109, 1164, 1270, 1412, 1429, 1442, 1478, 1512, 1582, 1607, 1645, 1678, 1811, 1819, 2150, 2215, 2292, 2339, 2443]
\end{quote}

The dominating set contains \textbf{35 nodes}, and the extracted subgraph remains within the 500-node limit.
\end{tcolorbox}
\end{figure*}

\begin{figure*}[t]
\begin{tcolorbox}[
    top=2pt,
    bottom=2pt,
    left=4pt,
    right=4pt,
    halign title=center,
    title={Graph Machine Learning Case Study}
]
\textbf{User Question}:Use graph machine learning algorithms to train a graph machine learning model, and use the graph machine learning model for inference. Predict whether there is a link between Node 111 and Node 350.

\medskip
\textbf{Agent Output}: Let me address this link prediction task. I will first inspect the basic properties of the dataset and determine whether an edge already exists between Node 111 and Node 350.

The initial inspection shows that Node 111 has degree 0 and is therefore an isolated node, whereas Node 350 has degree 7. No edge currently exists between the two nodes, and they have no common neighbors.

I will now train a link prediction model. The model training is complete. I will create the query edge and perform inference. After obtaining the prediction, I will also apply several graph-theoretic methods for cross-validation.

\medskip
\textbf{Dataset Information:}

\begin{center}
\begin{tabular}{ll}
\hline
Property & Value \\
\hline
Dataset & Cora citation network \\
Number of nodes & 2,708 \\
Number of edges & 5,277 (undirected) \\
Feature dimension & 1,433 \\
Degree of Node 111 & 0 (isolated node) \\
Degree of Node 350 & 7 \\
\hline
\end{tabular}
\end{center}

\medskip
\textbf{Graph-theoretic Link Prediction Results for Nodes 111 and 350:}

\begin{center}
\begin{tabular}{ll}
\hline
Method & Result \\
\hline
Existing edge & No \\
Existing path & No; Node 111 is isolated \\
Jaccard coefficient & 0.0 \\
Adamic--Adar index & 0.0 \\
Resource allocation index & 0.0 \\
\hline
\end{tabular}
\end{center}

These conventional heuristic methods all return zero because Node 111 has no neighbors.

\medskip
\textbf{GCN Link Prediction Results:}

\begin{center}
\begin{tabular}{ll}
\hline
Property & Value \\
\hline
Model architecture & GCN \\
Number of hidden channels & 64 \\
Number of training epochs & 200 \\
Predicted probability & 0.8407 (84.07\%) \\
Predicted result & A link should exist \\
\hline
\end{tabular}
\end{center}

\medskip
\textbf{Analysis:}

The GCN model assigns a probability of 84.07\% to the existence of a link between Node 111 and Node 350.

This result is notable because Node 111 is an isolated node with degree 0, causing conventional graph-theoretic heuristics, such as the Jaccard coefficient and Adamic--Adar index, to return zero. In contrast, the GCN-based link prediction model can still use the content features of the nodes, represented by 1,433-dimensional bag-of-words vectors. Although Node 111 has no neighboring nodes from which to aggregate information, its own content features can still contribute to its learned representation. The model therefore determines that the two papers are sufficiently similar in the embedding space to warrant a citation link, even though no such edge currently exists in the observed graph.
\end{tcolorbox}
\end{figure*}

\begin{figure*}[t]
\begin{tcolorbox}[
    top=2pt,
    bottom=2pt,
    left=4pt,
    right=4pt,
    halign title=center,
    title={Graph Open-ended Question Answering Case Study}
]
\footnotesize

\textbf{User Question}: 
As part of a project to map the intellectual landscape of our flagship conferences and journals over the past two decades, we need to trace the shortest citation chain of knowledge evolution linking Paper A (Node ID 77) to Paper B (Node ID 129). To reduce distortion from overly dominant hubs, first remove from the entire citation network those highly connected papers that receive and send many citations, thereby exposing the underlying structural patterns.

Within the resulting network, reconstruct the shortest path between Paper A and Paper B. We will then assess how tightly the direct neighbors of Paper A cite one another to quantify the cohesion of its immediate intellectual community. Building on this local view, identify the frontier papers exactly two citation steps away from Paper A---papers that lie on the perimeter of its influence but are not directly adjacent to it. Finally, within the 3-hop neighborhood of Paper A, locate the critical connector papers whose removal would fragment the local citation structure, revealing the works that play an essential linking role.

Please perform this analysis and output the final results in JSON format. If multiple candidate papers are tied under the relevant structural criterion, select the one with the smallest numerical Node ID.

\medskip
\textbf{Agent Output}: 
Now I have the shortest path. Let me run the remaining analyses. I will need to write some custom scripts since some operations require precise logic not covered by a single tool.

All results have been verified. Here are the complete results of the Cora citation network analysis:

\medskip
\noindent
\begin{minipage}{\linewidth}
\ttfamily\footnotesize
\raggedright
\{ \\
\hspace*{1em}"shortest\_path\_A\_to\_B": [77, 337, 346, 163, 129], \\
\hspace*{1em}"shortest\_path\_length": 4, \\
\hspace*{1em}"neighbor\_clustering\_coefficient": 0.0, \\
\hspace*{1em}"neighbor\_clustering\_note": "Node 77 has only one neighbor (337); clustering among neighbors is trivially 0.0 (insufficient pairs)", \\
\hspace*{1em}"two\_hop\_frontier": [143, 346, 363], \\
\hspace*{1em}"pruned\_shortest\_path\_A\_to\_B": null, \\
\hspace*{1em}"pruned\_shortest\_path\_length": null, \\
\hspace*{1em}"pruned\_removal\_note": "Top 5\% (136 nodes, min degree >= 18) removed. Node 129 (degree=30, rank=48) was among removed nodes.", \\
\hspace*{1em}"articulation\_points\_in\_3hop\_neighborhood": [346, 143, 337], \\
\hspace*{1em}"critical\_connector\_paper": 143, \\
\hspace*{1em}"critical\_connector\_impact": "Removal fragments 3-hop neighborhood into 3 components", \\
\hspace*{1em}"tiebreaker\_rationale": "Nodes 143 and 346 both create 3 components; 143 chosen per smallest-ID tiebreaker" \\
\}
\end{minipage}

\medskip
\textbf{What each result means:}

\smallskip
\textbf{Shortest knowledge chain (77 $\rightarrow$ 129):}
\texttt{77 $\rightarrow$ 337 $\rightarrow$ 346 $\rightarrow$ 163 $\rightarrow$ 129}, corresponding to four citation hops. Papers 77 and 337 cite each other; Paper 337 cites Paper 346; Paper 346 cites Paper 163, the network's highest-degree hub with degree 336; and Paper 163 cites Paper 129.

\smallskip
\textbf{Neighbor cohesion:}
Paper 77 has exactly one direct neighbor, Paper 337, with which it has a mutual citation. With only one neighbor, there are no pairs of neighboring papers that can be connected, so the clustering coefficient is trivially 0.0. This indicates that Paper 77 lies on the sparse periphery of the network and has almost no immediate intellectual community.

\smallskip
\textbf{2-hop frontier:}
\texttt{[143, 346, 363]}. These three papers are exactly two citation steps from Paper 77 and have no direct citation edge to or from it. They therefore form the outer boundary of Paper 77's influence.

\smallskip
\textbf{Top-5\% filtering:}
Removing the 136 most highly connected papers, whose total degrees are at least 18, eliminates Paper 129 itself, which has degree 30 and rank 48. Its hub neighbor, Paper 163, which has degree 336 and rank 1, is also removed. Consequently, no path from Paper 77 to Paper 129 exists in the pruned graph because the target paper itself has been removed.

\smallskip
\textbf{Critical connectors in the 3-hop neighborhood:}
The nine-node local subgraph contains three articulation points: \texttt{337}, \texttt{143}, and \texttt{346}. Removing either Paper 143 or Paper 346 divides the neighborhood into three components. Paper 143 is selected according to the smallest-ID tie-breaking rule. Its removal disconnects Papers 201 and 180 from the rest of the local structure, making it the essential linking work.

\end{tcolorbox}
\end{figure*}

%% file: graph_statistics.tex
\begin{table}[h]
\centering
\caption{Statistics of the 13 real-world datasets included in GABench, grouped by graph type.} %
\label{tab:graph_statistics} %
\resizebox{\columnwidth}{!}{%
\begin{tabular}{cccccc} %
\toprule
\textbf{Type} & \textbf{Name} & \textbf{\#Graphs} & \textbf{\#Nodes} & \textbf{\#Edges} & \textbf{Domain} \\
\midrule
\multirow{6}{*}{\textbf{TAG}} & Reddit     & 1       & 33,434    & 302,876     & Social         \\
                              & Instagram  & 1       & 11,339    & 155,349     & Social        \\
                              & History    & 1       & 41,551    & 503,180     & E-commerce    \\
                              & Product    & 1       & 2,449,029 & 123,618,192 & E-commerce    \\
                              & Arxiv      & 1       & 169,343   & 2,315,598   & CS Citation   \\
                              & Cora       & 1       & 2,708     & 10,556      & CS Citation    \\
\midrule
\multirow{5}{*}{\textbf{NAG}} & DGraph-Fin & 1       & 3,700,550 & 4,300,999 & Finance        \\
                              & Pokec      & 1       & 66,569    & 729,129   & Social        \\
                              & NBA        & 1       & 403       & 16,570    & Social         \\
                              & Credit     & 1       & 30,000    & 137,377   & Finance        \\
                              & Bail       & 1       & 18,876    & 311,870   & Justice        \\
\midrule
\multirow{2}{*}{\textbf{TPG}} & HIV        & 41,127  & 26        & 55          & Molecule      \\
                              & PCBA       & 437,092 & 26        & 56          & Molecule     \\
\bottomrule
\end{tabular}%
}
\end{table}

%% file: model_temp.tex
\begin{table}[h]
\centering
\caption{Roles and decoding configurations of the LLMs used in GABench.}
\label{tab:model_temp}
\small
\setlength{\tabcolsep}{4pt}
\renewcommand{\arraystretch}{1.05}
\begin{tabularx}{\columnwidth}{@{}
>{\centering\arraybackslash}X
>{\centering\arraybackslash}X
c c@{}}
\toprule
\textbf{Model}
& \textbf{Role}
& \textbf{Temperature}
& \textbf{Top-$p$} \\
\midrule
Qwen3-235B          & Evaluated agent    & 0.60 & 0.95 \\
GLM-4.5-flash       & Evaluated agent    & 0.70 & 0.95 \\
GLM-5-turbo               & Evaluated agent    & 1.00 & 1.00 \\
DeepSeek-V4-pro     & Evaluated agent    & 1.00 & 1.00 \\
GPT-5-mini          & Evaluated agent    & 1.00 & 1.00 \\
Gemini-3-flash      & Evaluated agent    & 1.00 & 0.95 \\
\midrule
GPT-4o-mini         & Question generator & 1.00 & 1.00 \\
DeepSeek-V4-Flash   & Judge              & 1.00 & 1.00 \\
\bottomrule
\end{tabularx}
\end{table}

%% file: token_retrieval.tex
\begin{table*}[t] 
\centering 
\caption{Input and output token consumption of each LLM agent on graph retrieval tasks, grouped by graph type and by node-, edge-, and graph-level subtasks. The rightmost column reports the mean token count across all subtask settings.} 
\label{tab:token_retrieval}
\setlength{\tabcolsep}{3.5pt} %
\resizebox{\textwidth}{!}{%
\begin{tabular}{c l
  >{\centering\arraybackslash}m{1.05cm}
  >{\centering\arraybackslash\hspace*{3pt}}m{1.05cm}
  >{\centering\arraybackslash}m{1.05cm}
  >{\centering\arraybackslash}m{1.05cm}
  >{\centering\arraybackslash\hspace*{2pt}}m{1.05cm}
  >{\centering\arraybackslash}m{1.05cm}
  >{\centering\arraybackslash}m{1.05cm}
  >{\centering\arraybackslash\hspace*{2pt}}m{1.05cm}
  >{\centering\arraybackslash}m{1.05cm}
  c} 
\toprule 
\multirow{2}{*}{\textbf{Type}} & \multirow{2}{*}{\textbf{Model}} & 
\multicolumn{3}{c}{\textbf{Numerical-Attribute Graph}} & 
\multicolumn{3}{c}{\textbf{Text-Attributed Graph}} & 
\multicolumn{3}{c}{\textbf{Text-Paired Graph}} & 
\multirow{2}{*}{\textbf{Avg. Tokens}} \\ 
\cmidrule(lr){3-5} \cmidrule(lr){6-8} \cmidrule(lr){9-11} 
& & \textbf{Node} & \textbf{Edge} & \textbf{Graph} & \textbf{Node} & \textbf{Edge} & \textbf{Graph} & \textbf{Node} & \textbf{Edge} & \textbf{Graph} & \\ 
\midrule

% ==================== INPUT TOKEN ====================
\multirow{6}{*}{\textbf{Input}} 
& Qwen3-235B      & 158,689 & 108,177 & 151,800 & 111,488 & 144,227 & 205,674 & 242,451 & 275,229 & 336,428 & 192,685 \\ 
& GLM-4.5-flash   & 16,188  & 9,784   & 14,318  & 21,904  & 16,202  & 56,081  & 20,309  & 38,373  & 31,959  & 31,752  \\ 
& GLM-5-turbo           & 17,523  & 12,899  & 64,186  & 19,426  & 16,173  & 29,523  & 23,630  & 26,557  & 20,358  & 24,528  \\ 
& DeepSeek-V4-pro & 26,791  & 59,795  & 59,410  & 35,844  & 32,610  & 34,899  & 32,060  & 28,585  & 33,206  & 35,179  \\ 
& GPT-5-mini      & 65,398  & 67,846  & 102,916 & 47,022  & 33,033  & 52,420  & 153,314 & 166,969 & 171,230 & 81,791  \\ 
& Gemini-3-flash  & 141,950 & 129,514 & 268,261 & 210,166 & 198,400 & 232,260 & 510,192 & 501,990 & 440,133 & 292,541 \\ 
\midrule

% ==================== OUTPUT TOKEN ====================
\multirow{6}{*}{\textbf{Output}} 
& Qwen3-235B      & 14,139  & 10,551  & 12,116  & 9,471   & 10,208  & 12,388  & 13,675  & 17,718  & 16,663  & 12,301  \\ 
& GLM-4.5-flash   & 830    & 870    & 950    & 2,052   & 2,517   & 2,227   & 3,409   & 2,398   & 2,201   & 2,297   \\ 
& GLM-5-turbo           & 1,525   & 1,658   & 10,511  & 1,516   & 1,363   & 2,879   & 2,446   & 2,739   & 2,669   & 2,724   \\ 
& DeepSeek-V4-pro & 3,909   & 4,165   & 7,685   & 3,356   & 2,960   & 4,340   & 2,969   & 4,289   & 4,800   & 3,944   \\ 
& GPT-5-mini      & 4,315   & 4,892   & 6,952   & 3,513   & 3,280   & 5,237   & 10,339  & 8,875   & 9,984   & 5,737   \\ 
& Gemini-3-flash  & 4,655   & 4,443   & 8,426   & 4,210   & 3,958   & 4,870   & 4,189   & 4,052   & 3,758   & 4,619   \\ 
\bottomrule 
\end{tabular}% 
}
\end{table*}

%% file: token_theory.tex
\begin{table*}[t] 
\centering 
\caption{Input and output token consumption of each LLM agent on graph theory tasks, grouped by graph type and by node-, edge-, and graph-level subtasks. The rightmost column reports the mean token count across all subtask settings.} 
\label{tab:graph_theory_tokens_single}
\setlength{\tabcolsep}{3pt} %
\resizebox{\textwidth}{!}{%
\begin{tabular}{c l
  >{\centering\arraybackslash}m{1.05cm}
  >{\centering\arraybackslash\hspace*{3pt}}m{1.05cm}
  >{\centering\arraybackslash}m{1.05cm}
  >{\centering\arraybackslash}m{1.05cm}
  >{\centering\arraybackslash\hspace*{2pt}}m{1.05cm}
  >{\centering\arraybackslash}m{1.05cm}
  >{\centering\arraybackslash}m{1.05cm}
  >{\centering\arraybackslash\hspace*{2pt}}m{1.05cm}
  >{\centering\arraybackslash}m{1.05cm}
  c} 
\toprule 
\multirow{2}{*}{\textbf{Type}} & \multirow{2}{*}{\textbf{Model}} & 
\multicolumn{3}{c}{\textbf{Numerical-Attribute Graph}} & 
\multicolumn{3}{c}{\textbf{Text-Attributed Graph}} & 
\multicolumn{3}{c}{\textbf{Text-Paired Graph}} & 
\multirow{2}{*}{\textbf{Avg. Tokens} }\\ 
\cmidrule(lr){3-5} \cmidrule(lr){6-8} \cmidrule(lr){9-11} 
& & \textbf{Node} & \textbf{Edge} & \textbf{Graph} & \textbf{Node} & \textbf{Edge} & \textbf{Graph} & \textbf{Node} & \textbf{Edge} & \textbf{Graph} & \\ 
\midrule

% ==================== INPUT TOKEN ====================
\multirow{6}{*}{\textbf{Input}} 
& Qwen3-235B      & 240,607 & 195,130 & 142,826 & 179,124 & 150,012 & 124,331 & 201,444 & 188,860 & 321,302 & 181,338 \\ 
& GLM-4.5-flash   & 12,284  & 11,812  & 29,278  & 17,088  & 29,598  & 22,304  & 8,776   & 36,374  & 30,919  & 24,872  \\ 
& GLM-5-turbo           & 18,815  & 31,241  & 15,910  & 16,307  & 16,745  & 18,464  & 18,892  & 16,330  & 28,619  & 19,455  \\ 
& DeepSeek-V4-pro & 31,240  & 40,526  & 32,401  & 33,680  & 35,294  & 32,583  & 33,560  & 31,174  & 32,757  & 33,746  \\ 
& GPT-5-mini      & 64,789  & 42,898  & 86,857  & 29,490  & 30,152  & 31,432  & 142,415 & 151,882 & 160,786 & 82,300  \\ 
& Gemini-3-flash  & 162,167 & 131,947 & 123,438 & 206,555 & 223,936 & 211,372 & 493,437 & 499,376 & 483,418 & 258,992 \\ 
\midrule

% ==================== OUTPUT TOKEN ====================
\multirow{6}{*}{\textbf{Output}} 
& Qwen3-235B      & 14,859  & 17,064  & 11,778  & 13,990  & 13,539  & 11,241  & 15,973  & 13,528  & 18,513  & 13,960  \\ 
& GLM-4.5-flash   & 1,453   & 1,788   & 2,273   & 2,097   & 3,066   & 2,161   & 3,567   & 4,241   & 3,839   & 2,620   \\ 
& GLM-5-turbo           & 1,865   & 2,184   & 6,173   & 1,364   & 2,244   & 1,930   & 2,148   & 2,552   & 2,890   & 2,418   \\ 
& DeepSeek-V4-pro & 5,069   & 6,069   & 4,743   & 3,446   & 4,250   & 3,453   & 3,546   & 4,094   & 4,654   & 4,248   \\ 
& GPT-5-mini      & 5,081   & 5,517   & 6,046   & 3,670   & 4,880   & 4,157   & 12,207  & 13,675  & 10,425  & 6,446   \\ 
& Gemini-3-flash  & 5,632   & 8,003   & 4,623   & 4,145   & 5,193   & 4,257   & 4,072   & 4,315   & 3,932   & 4,765   \\ 
\bottomrule 
\end{tabular}% 
}
\end{table*}

%% file: token_ml.tex
\begin{table*}[t] 
\centering 
\caption{Input and output token consumption of each LLM agent on graph machine learning tasks. Results are grouped into node classification (NC), link prediction (LP), and graph classification (GC) across the applicable graph types; the rightmost column reports the mean token count across these settings.} 
\label{tab:graph_ml_tokens}
\large
\setlength{\tabcolsep}{4.5pt} %
\resizebox{\textwidth}{!}{%
\begin{tabular}{c l cc cc c c} 
\toprule 
\multirow{2}{*}{\textbf{Type}} & \multirow{2}{*}{\textbf{Model}} & 
\multicolumn{2}{c}{\textbf{Numerical-Attribute Graph}} & 
\multicolumn{2}{c}{\textbf{Text-Attributed Graph}} & 
\multicolumn{1}{c}{\textbf{Text-Paired Graph}} & 
\multirow{2}{*}{\textbf{Avg. Tokens}} \\ 
\cmidrule(lr){3-4} \cmidrule(lr){5-6} \cmidrule(lr){7-7} 
& & {\normalsize\textbf{Node Classification}} & {\normalsize\textbf{Link Prediction}} & {\normalsize\textbf{Node Classification}} & {\normalsize\textbf{Link Prediction}} & {\normalsize\textbf{Graph Classification}} & \\ 
\midrule

\multirow{6}{*}{\textbf{Input}} 
& Qwen3-235B      & 223,618 & 241,032 & 251,132 & 334,030 & 463,736 & 324,905 \\ 
& GLM-4.5-flash   & 89,393  & 67,220  & 85,497  & 70,829  & 14,826  & 70,877  \\ 
& GLM-5-turbo           & 116,609 & 29,022  & 75,493  & 31,952  & 54,080  & 61,431  \\ 
& DeepSeek-V4-pro & 106,296 & 33,126  & 81,084  & 32,027  & 43,201  & 56,786  \\ 
& GPT-5-mini      & 256,234 & 82,732  & 192,134 & 90,702  & 163,832 & 149,897 \\ 
& Gemini-3-flash  & 485,526 & 305,596 & 557,985 & 328,847 & 668,516 & 492,259 \\ 
\midrule
\multirow{6}{*}{\textbf{Output}} 
& Qwen3-235B      & 12,552  & 11,677  & 10,939  & 12,281  & 17,860  & 13,287  \\ 
& GLM-4.5-flash   & 3,588   & 2,587   & 4,094   & 4,126   & 8,817   & 5,043   \\ 
& GLM-5-turbo           & 2,968   & 2,742   & 5,708   & 2,879   & 6,708   & 4,201   \\ 
& DeepSeek-V4-pro & 6,062   & 7,073   & 7,049   & 6,691   & 8,591   & 7,051   \\ 
& GPT-5-mini      & 5,719   & 7,930   & 7,154   & 7,401   & 13,777  & 8,489   \\ 
& Gemini-3-flash  & 4,130   & 2,896   & 3,838   & 2,777   & 6,066   & 4,055   \\ 
\bottomrule 
\end{tabular}% 
}
\end{table*}

%% file: token_openend.tex
\begin{table*}[t] 
\centering 
\caption{Input and output token consumption of each LLM agent on open-ended graph question-answering tasks, grouped into insight and prediction subtasks across the three graph types. The two rightmost columns report the mean token counts for the corresponding subtask across graph types.} 
\label{tab:token_open}
\setlength{\tabcolsep}{3.2pt} %
\resizebox{\textwidth}{!}{%
\begin{tabular}{c l
  *{4}{
    >{\centering\arraybackslash\hspace*{-2pt}}m{1.7cm}
    >{\centering\arraybackslash\hspace*{2pt}}m{1.7cm}
  }} 
\toprule 
\multirow{2}{*}{\textbf{Type}} & \multirow{2}{*}{\textbf{Model}} & 
\multicolumn{2}{c}{\textbf{Numerical-Attribute Graph}} & 
\multicolumn{2}{c}{\textbf{Text-Attributed Graph}} & 
\multicolumn{2}{c}{\textbf{Text-Paired Graph}} & 
\multicolumn{2}{c}{\textbf{Avg. Tokens}} \\ 
\cmidrule(lr){3-4} \cmidrule(lr){5-6} \cmidrule(lr){7-8} \cmidrule(lr){9-10}
& & \textbf{Insight} & \textbf{Prediction} & \textbf{Insight} & \textbf{Prediction} & \textbf{Insight} & \textbf{Prediction} & \textbf{Insight} & \textbf{Prediction} \\ 
\midrule

% ==================== INPUT TOKEN ====================
\multirow{6}{*}{\textbf{Input}} 
& Qwen3-235B      & 174,554 & 210,302 & 178,755 & 189,678 & 208,716 & 196,545 & 187,342 & 197,521 \\ 
& GLM-4.5-flash   & 36,752  & 61,559  & 51,577  & 62,041  & 46,757  & 50,709  & 48,409  & 59,643  \\ 
& GLM-5-turbo           & 43,488  & 76,421  & 50,633  & 93,702  & 25,281  & 49,729  & 44,054  & 80,214  \\ 
& DeepSeek-V4-pro & 43,375  & 68,677  & 52,952  & 87,789  & 41,361  & 39,090  & 47,990  & 73,718  \\ 
& GPT-5-mini      & 223,912 & 265,024 & 208,977 & 258,312 & 236,340 & 115,361 & 219,626 & 228,204 \\ 
& Gemini-3-flash  & 327,255 & 389,068 & 481,337 & 654,731 & 605,432 & 717,627 & 457,970 & 587,471 \\ 
\midrule

% ==================== OUTPUT TOKEN ====================
\multirow{6}{*}{\textbf{Output}} 
& Qwen3-235B      & 19,541  & 12,121  & 9,913   & 10,172  & 9,765   & 8,889   & 10,117  & 10,512  \\ 
& GLM-4.5-flash   & 3,690   & 6,027   & 6,170   & 7,131   & 4,324   & 7,833   & 5,464   & 6,971   \\ 
& GLM-5-turbo           & 11,539  & 13,724  & 12,441  & 15,214  & 8,840   & 11,249  & 11,541  & 14,021  \\ 
& DeepSeek-V4-pro & 12,752  & 17,317  & 14,158  & 18,432  & 13,204  & 10,246  & 13,567  & 16,754  \\ 
& GPT-5-mini      & 20,345  & 23,100  & 18,938  & 23,061  & 13,258  & 14,122  & 17,983  & 21,060  \\ 
& Gemini-3-flash  & 16,370  & 13,631  & 13,516  & 11,022  & 10,556  & 7,194   & 13,811  & 11,039  \\ 
\bottomrule 
\end{tabular}% 
}
\end{table*}

%% file: tokenml_harness.tex
% Required packages:
% \usepackage{booktabs}
% \usepackage{multirow}
% \usepackage{graphicx}

\begin{table*}[t]
    \centering
    \caption{Input and output token usage across different agent harnesses on graph machine learning tasks.}
    \label{tab:harness_token_ml}
    \setlength{\tabcolsep}{5pt}
    \renewcommand{\arraystretch}{1.12}
    \resizebox{\textwidth}{!}{%
    \begin{tabular}{lrrrrrrrrrrrr}
        \toprule
        \multirow{3}{*}{Harness}
        & \multicolumn{4}{c}{Numerical-Attribute Graph}
        & \multicolumn{4}{c}{Text-Attributed Graph}
        & \multicolumn{2}{c}{Text-Paired Graph}
        & \multicolumn{2}{c}{Average Tokens} \\[-2pt]
        \cmidrule(lr){2-5}
        \cmidrule(lr){6-9}
        \cmidrule(lr){10-11}
        \cmidrule(lr){12-13}

        & \multicolumn{2}{c}{Node Classification}
        & \multicolumn{2}{c}{Link Prediction}
        & \multicolumn{2}{c}{Node Classification}
        & \multicolumn{2}{c}{Link Prediction}
        & \multicolumn{2}{c}{Graph Classification}
        & \multicolumn{2}{c}{Overall} \\[-1pt]
        \cmidrule(lr){2-3}
        \cmidrule(lr){4-5}
        \cmidrule(lr){6-7}
        \cmidrule(lr){8-9}
        \cmidrule(lr){10-11}
        \cmidrule(lr){12-13}

        & Input & Output
        & Input & Output
        & Input & Output
        & Input & Output
        & Input & Output
        & Input & Output \\[-3pt]
        \midrule

        Hermes
        & 25,112  & 5,157
        & 12,016  & 3,139
        & 37,606  & 3,478
        & 26,162  & 3,154
        & 48,079  & 5,292
        & 29,795  & 4,044 \\

        ClaudeCode
        & 10,668  & 4,176
        & 10,995  & 5,108
        & 18,153  & 4,183
        & 22,071  & 4,807
        & 27,081  & 5,857
        & 15,793  & 4,826 \\

        OpenClaw
        & 116,609 & 2,968
        & 29,022  & 2,742
        & 75,493  & 5,708
        & 31,952  & 2,879
        & 54,080  & 6,708
        & 61,431  & 4,201 \\

        \bottomrule
    \end{tabular}%
    }
\end{table*}

%% file: tokenopen_harness.tex
% Required packages:
% \usepackage{booktabs}
% \usepackage{multirow}
% \usepackage{graphicx}

\begin{table*}[t]
    \centering
    \caption{Input and output token usage of different agent harnesses on graph open-ended question answering tasks.}
    \label{tab:harness_token_open}
    \setlength{\tabcolsep}{4.5pt}
    \renewcommand{\arraystretch}{1.12}
    \resizebox{\textwidth}{!}{%
    \begin{tabular}{lrrrrrrrrrrrrrr}
        \toprule
        \multirow{3}{*}{Harness}
        & \multicolumn{4}{c}{Numerical-Attribute Graph}
        & \multicolumn{4}{c}{Text-Attributed Graph}
        & \multicolumn{4}{c}{Text-Paired Graph}
        & \multicolumn{2}{c}{Average Tokens} \\[-2pt]
        \cmidrule(lr){2-5}
        \cmidrule(lr){6-9}
        \cmidrule(lr){10-13}
        \cmidrule(lr){14-15}

        & \multicolumn{2}{c}{Insight}
        & \multicolumn{2}{c}{Prediction}
        & \multicolumn{2}{c}{Insight}
        & \multicolumn{2}{c}{Prediction}
        & \multicolumn{2}{c}{Insight}
        & \multicolumn{2}{c}{Prediction}
        & \multicolumn{2}{c}{Overall} \\[-1pt]
        \cmidrule(lr){2-3}
        \cmidrule(lr){4-5}
        \cmidrule(lr){6-7}
        \cmidrule(lr){8-9}
        \cmidrule(lr){10-11}
        \cmidrule(lr){12-13}
        \cmidrule(lr){14-15}

        & Input & Output
        & Input & Output
        & Input & Output
        & Input & Output
        & Input & Output
        & Input & Output
        & Input & Output \\[-3pt]
        \midrule

        Hermes
        & 25,541 & 8,130
        & 34,769 & 10,375
        & 28,896 & 7,879
        & 39,667 & 9,444
        & 35,604 & 7,580
        & 47,829 & 6,839
        & 35,384 & 8,374 \\

        ClaudeCode
        & 15,835 & 7,970
        & 16,075 & 9,799
        & 23,048 & 7,249
        & 21,818 & 8,179
        & 28,184 & 7,377
        & 32,134 & 7,440
        & 22,849 & 8,002 \\

        OpenClaw
        & 43,488 & 11,539
        & 76,421 & 13,724
        & 50,633 & 12,441
        & 93,702 & 15,214
        & 25,281 & 8,840
        & 49,729 & 11,249
        & 56,542 & 12,167 \\

        \bottomrule
    \end{tabular}%
    }
\end{table*}

%% file: reference.bib
@misc{qin2024nasbenchgraphbenchmarkinggraphneural,
      title={NAS-Bench-Graph: Benchmarking Graph Neural Architecture Search}, 
      author={Yijian Qin and Ziwei Zhang and Xin Wang and others},
      year={2024},
      eprint={2206.09166},
      archivePrefix={arXiv},
      primaryClass={cs.LG}
}

@misc{yang2024cragcomprehensiverag,
      title={CRAG -- Comprehensive RAG Benchmark}, 
      author={Xiao Yang and Kai Sun and Hao Xin and others},
      year={2024},
      eprint={2406.04744},
      archivePrefix={arXiv},
      primaryClass={cs.CL}
}

@misc{ruan2024identifyingriskslmagents,
      title={Identifying the Risks of LM Agents with an LM-Emulated Sandbox}, 
      author={Yangjun Ruan and Honghua Dong and Andrew Wang and others},
      year={2024},
      eprint={2309.15817},
      archivePrefix={arXiv},
      primaryClass={cs.AI}
}

@inproceedings{gu2024blade,
  title={Blade: Benchmarking language model agents for data-driven science},
  author={Gu, Ken and Shang, Ruoxi and Jiang, Ruien and others},
  booktitle={Findings of the Association for Computational Linguistics: EMNLP 2024},
  pages={13936--13971},
  year={2024}
}

@article{ma2026can,
  title={Can ai agents answer your data questions? a benchmark for data agents},
  author={Ma, Ruiying and Shankar, Shreya and Chen, Ruiqi and Lin, Yiming and Zeighami, Sepanta and Ghosh, Rajoshi and Gupta, Abhinav and Gupta, Anushrut and Gopal, Tanmai and Parameswaran, Aditya G},
  journal={arXiv preprint arXiv:2603.20576},
  year={2026}
}

@article{hu2024infiagent,
  title={Infiagent-dabench: Evaluating agents on data analysis tasks},
  author={Hu, Xueyu and Zhao, Ziyu and Wei, Shuang and Chai, Ziwei and Ma, Qianli and Wang, Guoyin and Wang, Xuwu and Su, Jing and Xu, Jingjing and Zhu, Ming and others},
  journal={arXiv preprint arXiv:2401.05507},
  year={2024}
}

@misc{huang2023dgraphlargescalefinancialdataset,
	title        = {DGraph: A Large-Scale Financial Dataset for Graph Anomaly Detection},
	author       = {Xuanwen Huang and Yang Yang and Yang Wang and others},
	year         = 2023,
	url          = {https://arxiv.org/abs/2207.03579},
	eprint       = {2207.03579},
	archivePrefix = {arXiv},
	primaryClass = {cs.SI}
}

@misc{chen2024textspacegraphfoundationmodels,
	title        = {Text-space Graph Foundation Models: Comprehensive Benchmarks and New Insights},
	author       = {Zhikai Chen and Haitao Mao and Jingzhe Liu and others},
	year         = 2024,
	eprint       = {2406.10727},
	archivePrefix = {arXiv},
	primaryClass = {cs.LG}
}

@article{liu2025graph,
  title={Graph foundation models: Concepts, opportunities and challenges},
  author={Liu, Jiawei and Yang, Cheng and Lu, Zhiyuan and Chen, Junze and Li, Yibo and Zhang, Mengmei and Bai, Ting and Fang, Yuan and Sun, Lichao and Yu, Philip S and others},
  journal={IEEE Transactions on Pattern Analysis and Machine Intelligence},
  year={2025},
  publisher={IEEE}
}

@misc{li2024glbenchcomprehensivebenchmarkgraph,
	title        = {GLBench: A Comprehensive Benchmark for Graph with Large Language Models},
	author       = {Yuhan Li and Peisong Wang and Xiao Zhu and others},
	year         = 2024,
	url          = {https://arxiv.org/abs/2407.07457},
	eprint       = {2407.07457},
	archivePrefix = {arXiv},
	primaryClass = {cs.LG}
}

@misc{dai2021saydiscriminationlearningfair,
	title        = {Say No to the Discrimination: Learning Fair Graph Neural Networks with Limited Sensitive Attribute Information},
	author       = {Enyan Dai and Suhang Wang},
	year         = 2021,
	url          = {https://arxiv.org/abs/2009.01454},
	eprint       = {2009.01454},
	archivePrefix = {arXiv},
	primaryClass = {cs.LG}
}

@inproceedings{Ma_2022,
	title        = {Learning Fair Node Representations with Graph Counterfactual Fairness},
	author       = {Ma, Jing and Guo, Ruocheng and Wan, Mengting and others},
	year         = 2022,
	month        = {Feb},
	booktitle    = {Proceedings of the Fifteenth ACM International Conference on Web Search and Data Mining},
	publisher    = {ACM},
	series       = {WSDM'22},
	pages        = {695--703},
	DOI          = {10.1145/3488560.3498391},
	url          = {http://dx.doi.org/10.1145/3488560.3498391},
	collection   = {WSDM'22}
}

@article{chen2024llaga,
	title        = {Llaga: Large language and graph assistant},
	author       = {Chen, Runjin and Zhao, Tong and Jaiswal, Ajay and others},
	year         = 2024,
	journal      = {arXiv preprint arXiv:2402.08170}
}

@inproceedings{zhang2024benchmarking,
	title        = {Benchmarking data science agents},
	author       = {Zhang, Yuge and Jiang, Qiyang and Han, Xingyu and others},
	year         = 2024,
	booktitle    = {Proceedings of the 62nd Annual Meeting of the Association for Computational Linguistics (Volume 1: Long Papers)},
	pages        = {5677--5700}
}

@article{sun2026lambda,
	title        = {Lambda: A large model based data agent},
	author       = {Sun, Maojun and Han, Ruijian and Jiang, Binyan and others},
	year         = 2026,
	journal      = {Journal of the American Statistical Association},
	publisher    = {Taylor \& Francis},
	volume       = 121,
	number       = 553,
	pages        = {1--13}
}

@inproceedings{hong2025data,
	title        = {Data interpreter: An llm agent for data science},
	author       = {Hong, Sirui and Lin, Yizhang and Liu, Bang and others},
	year         = 2025,
	booktitle    = {Findings of the Association for Computational Linguistics: ACL 2025},
	pages        = {19796--19821}
}

@inproceedings{perez2025llm,
	title        = {An LLM-based approach for insight generation in data analysis},
	author       = {P{\'e}rez, Alberto S{\'a}nchez and Boukhary, Alaa and Papotti, Paolo and others},
	year         = 2025,
	booktitle    = {Proceedings of the 2025 Conference of the Nations of the Americas Chapter of the Association for Computational Linguistics: Human Language Technologies (Volume 1: Long Papers)},
	pages        = {562--582}
}

@article{ma2024agentboard,
	title        = {Agentboard: An analytical evaluation board of multi-turn llm agents},
	author       = {Ma, Chang and Zhang, Junlei and Zhu, Zhihao and others},
	year         = 2024,
	journal      = {Advances in neural information processing systems},
	volume       = 37,
	pages        = {74325--74362}
}

@misc{hermes,
	title        = {{Hermes}},
	author       = {{Hermes Team}},
	year         = 2026,
	journal      = {GitHub repository},
	publisher    = {GitHub},
	howpublished = {\url{https://github.com/nousresearch/hermes-agent}}
}

@misc{claudecode,
	title        = {{Claude Code}},
	author       = {{Claude Code Team}},
	year         = 2026,
	journal      = {GitHub repository},
	publisher    = {GitHub},
	howpublished = {\url{https://github.com/anthropics/claude-code}}
}

@misc{openclaw,
	title        = {{OpenClaw}},
	author       = {{OpenClaw Team}},
	year         = 2026,
	journal      = {GitHub repository},
	publisher    = {GitHub},
	howpublished = {\url{https://github.com/openclaw/openclaw}}
}

@misc{ding2026wildclawbenchbenchmarkrealworldlonghorizon,
	title        = {WildClawBench: A Benchmark for Real-World, Long-Horizon Agent Evaluation},
	author       = {Shuangrui Ding and Xuanlang Dai and Long Xing and others},
	year         = 2026,
	url          = {https://arxiv.org/abs/2605.10912},
	eprint       = {2605.10912},
	archivePrefix = {arXiv},
	primaryClass = {cs.CL}
}

@misc{geminiteam2025gemini,
	title        = {Gemini: A Family of Highly Capable Multimodal Models},
	author       = {{Gemini Team}},
	year         = 2025,
	eprint       = {2312.11805},
	archivePrefix = {arXiv},
	primaryClass = {cs.CL}
}

@misc{deepseekai2026deepseekv4,
	title        = {DeepSeek-V4: Towards Highly Efficient Million-Token Context Intelligence},
	author       = {{DeepSeek-AI}},
	year         = 2026,
	eprint       = {2606.19348},
	archivePrefix = {arXiv},
	primaryClass = {cs.CL}
}

@misc{glm5team2026glm5,
	title        = {{GLM-5}: From Vibe Coding to Agentic Engineering},
	author       = {{GLM-5 Team}},
	year         = 2026,
	eprint       = {2602.15763},
	archivePrefix = {arXiv},
	primaryClass = {cs.LG}
}

@article{yao2022webshop,
	title        = {Webshop: Towards scalable real-world web interaction with grounded language agents},
	author       = {Yao, Shunyu and Chen, Howard and Yang, John and others},
	year         = 2022,
	journal      = {Advances in Neural Information Processing Systems},
	volume       = 35,
	pages        = {20744--20757}
}

@misc{merrill2026terminalbenchbenchmarkingagentshard,
	title        = {Terminal-Bench: Benchmarking Agents on Hard, Realistic Tasks in Command Line Interfaces},
	author       = {Mike A. Merrill and Alexander G. Shaw and others},
	year         = 2026,
	url          = {https://arxiv.org/abs/2601.11868},
	eprint       = {2601.11868},
	archivePrefix = {arXiv},
	primaryClass = {cs.SE}
}

@article{wei2026graphchain,
	title        = {Graphchain: Large language models for large-scale graph analysis via tool chaining},
	author       = {Wei, Chunyu and Hu, Wenji and Hao, Xingjia and others},
	year         = 2026,
	journal      = {Advances in Neural Information Processing Systems},
	volume       = 38,
	pages        = {7402--7436}
}

@book{gross2018graph,
	title        = {Graph theory and its applications},
	author       = {Gross, Jonathan L and Yellen, Jay and Anderson, Mark},
	year         = 2018,
	publisher    = {Chapman and Hall/CRC}
}

@article{peng2025graph,
	title        = {Graph retrieval-augmented generation: A survey},
	author       = {Peng, Boci and Zhu, Yun and Liu, Yongchao and others},
	year         = 2025,
	journal      = {ACM Transactions on Information Systems},
	publisher    = {ACM New York, NY},
	volume       = 44,
	number       = 2,
	pages        = {1--52}
}

@article{bonald2020scikit,
	title        = {Scikit-network: Graph analysis in python},
	author       = {Bonald, Thomas and De Lara, Nathan and Lutz, Quentin and others},
	year         = 2020,
	journal      = {Journal of Machine Learning Research},
	volume       = 21,
	number       = 185,
	pages        = {1--6}
}

@article{perret2019higra,
	title        = {Higra: Hierarchical graph analysis},
	author       = {Perret, Benjamin and Chierchia, Giovanni and Cousty, Jean and others},
	year         = 2019,
	journal      = {SoftwareX},
	publisher    = {Elsevier},
	volume       = 10,
	pages        = 100335
}

@book{easley2010networks,
	title        = {Networks, crowds, and markets: Reasoning about a highly connected world},
	author       = {Easley, David and Kleinberg, Jon and others},
	year         = 2010,
	publisher    = {Cambridge university press Cambridge},
	volume       = 1
}

@misc{gao2025mcpradarmultidimensionalbenchmarkevaluating,
	title        = {MCP-RADAR: A Multi-Dimensional Benchmark for Evaluating Tool Use Capabilities in Large Language Models},
	author       = {Xuanqi Gao and Siyi Xie and Juan Zhai and others},
	year         = 2025,
	url          = {https://arxiv.org/abs/2505.16700},
	eprint       = {2505.16700},
	archivePrefix = {arXiv},
	primaryClass = {cs.AI}
}

@misc{wang2025mcpbenchbenchmarkingtoolusingllm,
	title        = {MCP-Bench: Benchmarking Tool-Using LLM Agents with Complex Real-World Tasks via MCP Servers},
	author       = {Zhenting Wang and Qi Chang and Hemani Patel and others},
	year         = 2025,
	url          = {https://arxiv.org/abs/2508.20453},
	eprint       = {2508.20453},
	archivePrefix = {arXiv},
	primaryClass = {cs.CL}
}

@article{DBLP:journals/corr/abs-2602-06319,
	title        = {Exposing Weaknesses of Large Reasoning Models through Graph Algorithm Problems},
	author       = {Qifan Zhang and Jianhao Ruan and Aochuan Chen and Kang Zeng and Nuo Chen and Jing Tang and Jia Li},
	year         = 2026,
	journal      = {CoRR},
	volume       = {abs/2602.06319},
	doi          = {10.48550/ARXIV.2602.06319},
	url          = {https://doi.org/10.48550/arXiv.2602.06319},
	eprinttype   = {arXiv},
	eprint       = {2602.06319},
	bibsource    = {dblp computer science bibliography, https://dblp.org}
}

@inproceedings{DBLP:conf/nips/LiCCLSLQW000Y24,
	title        = {Can Large Language Models Analyze Graphs like Professionals? {A} Benchmark, Datasets and Models},
	author       = {Xin Li and Weize Chen and Qizhi Chu and others},
	year         = 2024,
	booktitle    = {Advances in Neural Information Processing Systems 37: Annual Conference on Neural Information Processing Systems 2024, NeurIPS 2024, Vancouver, BC, Canada, December 10 - 15, 2024}
}

@article{tang2024grapharena,
	title        = {Grapharena: Benchmarking large language models on graph computational problems},
	author       = {Tang, Jianheng and Zhang, Qifan and Li, Yuhan and others},
	year         = 2024,
	journal      = {arXiv e-prints},
	pages        = {arXiv--2407}
}

@article{yuan2024gracore,
	title        = {Gracore: Benchmarking graph comprehension and complex reasoning in large language models},
	author       = {Yuan, Zike and Liu, Ming and Wang, Hui and others},
	year         = 2024,
	journal      = {arXiv preprint arXiv:2407.02936}
}

@article{xu2025graphomni,
	title        = {GraphOmni: A Comprehensive and Extendable Benchmark Framework for Large Language Models on Graph-theoretic Tasks},
	author       = {Xu, Hao and Jian, Xiangru and Zhao, Xinjian and others},
	year         = 2025,
	journal      = {arXiv preprint arXiv:2504.12764}
}

@article{luo2024graphinstruct,
	title        = {Graphinstruct: Empowering large language models with graph understanding and reasoning capability},
	author       = {Luo, Zihan and Song, Xiran and Huang, Hong and others},
	year         = 2024,
	journal      = {arXiv preprint arXiv:2403.04483}
}

@article{wang2023can,
	title        = {Can language models solve graph problems in natural language?},
	author       = {Wang, Heng and Feng, Shangbin and He, Tianxing and others},
	year         = 2023,
	journal      = {Advances in Neural Information Processing Systems},
	volume       = 36,
	pages        = {30840--30861}
}

@inproceedings{zhang2024llm4dyg,
	title        = {LLM4DyG: Can large language models solve spatial-temporal problems on dynamic graphs?},
	author       = {Zhang, Zeyang and Wang, Xin and Zhang, Ziwei and others},
	year         = 2024,
	booktitle    = {Proceedings of the 30th ACM SIGKDD Conference on Knowledge Discovery and Data Mining},
	pages        = {4350--4361}
}

@inproceedings{chen2024graphwiz,
	title        = {Graphwiz: An instruction-following language model for graph computational problems},
	author       = {Chen, Nuo and Li, Yuhan and Tang, Jianheng and others},
	year         = 2024,
	booktitle    = {Proceedings of the 30th ACM SIGKDD Conference on Knowledge Discovery and Data Mining},
	pages        = {353--364}
}

@article{guo2023gpt4graph,
	title        = {Gpt4graph: Can large language models understand graph structured data? an empirical evaluation and benchmarking},
	author       = {Guo, Jiayan and Du, Lun and Liu, Hengyu and others},
	year         = 2023,
	journal      = {arXiv preprint arXiv:2305.15066}
}

@article{openai2025gpt5,
	title        = {Introducing GPT-5},
	author       = {{OpenAI}},
	year         = 2025,
	month        = {August},
	journal      = {OpenAI},
	note         = {Accessed: 2025-08-07}
}

@article{zheng2025livecodebench,
	title        = {LiveCodeBench Pro: How Do Olympiad Medalists Judge LLMs in Competitive Programming?},
	author       = {Zheng, Zihan and Cheng, Zerui and Shen, Zeyu and others},
	year         = 2025,
	journal      = {arXiv preprint arXiv:2506.11928}
}

@inproceedings{chen2024link,
	title        = {Link recommendation to augment influence diffusion with provable guarantees},
	author       = {Chen, Xiaolong and Song, Yifan and Tang, Jing},
	year         = 2024,
	booktitle    = {Proceedings of the ACM Web Conference 2024},
	pages        = {2509--2518}
}

@article{jimenez2023swe,
	title        = {Swe-bench: Can language models resolve real-world github issues?},
	author       = {Jimenez, Carlos E and Yang, John and Wettig, Alexander and others},
	year         = 2023,
	journal      = {arXiv preprint arXiv:2310.06770}
}

@inproceedings{Liu-ICLR2024,
	title        = {Agent{B}ench: Evaluating {LLM}s as Agents},
	author       = {Xiao Liu and Hao Yu and Hanchen Zhang and others},
	year         = 2024,
	month        = {Jan.},
	booktitle    = {International Conference on Learning Representations (ICLR)},
	url          = {https://openreview.net/forum?id=zAdUB0aCTQ}
}

@inproceedings{Xie-ICML2024,
	title        = {{TravelPlanner}: A Benchmark for Real-World Planning with Language Agents},
	author       = {Xie, Jian and Zhang, Kai and Chen, Jiangjie and others},
	year         = 2024,
	month        = {Jan.},
	booktitle    = {International Conference on Machine Learning (ICML)},
}

@inproceedings{Zhou-ICLR2024,
	title        = {WebArena: A Realistic Web Environment for Building Autonomous Agents},
	author       = {Shuyan Zhou and Frank F. Xu and Hao Zhu and others},
	year         = 2024,
	month        = {Jan.},
	booktitle    = {International Conference on Learning Representations (ICLR)},
}

@inproceedings{Yao-ICLR2025,
	title        = {$\tau$-{B}ench: Evaluating Tool-Augmented Language Agents through Human-in-the-Loop Collaboration},
	author       = {Shunyu Yao and Noah Shinn and Pedram Razavi and others},
	year         = 2025,
	month        = {Jan.},
	booktitle    = {International Conference on Learning Representations (ICLR)}
}

@inproceedings{Patil-ICML2025,
	title        = {The Berkeley Function Calling Leaderboard ({BFCL}): From Tool Use to Agentic Evaluation of Large Language Models},
	author       = {Shishir G Patil and Huanzhi Mao and Fanjia Yan and others},
	year         = 2025,
	month        = {May},
	booktitle    = {International Conference on Machine Learning (ICML)},
	url          = {https://openreview.net/forum?id=2GmDdhBdDk}
}

@article{Deng-NeurIPS2023,
	title        = {{Mind2Web}: Towards a generalist agent for the web},
	author       = {Deng, Xiang and Gu, Yu and Zheng, Boyuan and others},
	year         = 2023,
	month        = {Sept.},
	journal      = {Advances in Neural Information Processing Systems (NeurIPS)},
	volume       = 36,
	pages        = {28091--28114},
}

@article{Koh-arXiv2024,
	title        = {Visualwebarena: Evaluating multimodal agents on realistic visual web tasks},
	author       = {Koh, Jing Yu and Lo, Robert and Jang, Lawrence and others},
	year         = 2024,
	journal      = {arXiv preprint}
}

@article{Yan-arXiv2025,
	title        = {{MCPWorld}: A Unified Benchmarking Testbed for {API}, {GUI}, and Hybrid Computer Use Agents},
	author       = {Yan, Yunhe and Wang, Shihe and Du, Jiajun and others},
	year         = 2025,
	journal      = {arXiv preprint},
	url          = {https://arxiv.org/abs/2506.07672}
}

@article{Liu-arXiv2025,
	title        = {{MCPEval}: Automatic {MCP}-Based Deep Evaluation for AI Agent Models},
	author       = {Zhiwei Liu and Jielin Qiu and Shiyu Wang and others},
	year         = 2025,
	journal      = {arXiv preprint}
}

@article{yang2025qwen3,
	title        = {Qwen3 technical report},
	author       = {Yang, An and Li, Anfeng and Yang, Baosong and others},
	year         = 2025,
	journal      = {arXiv preprint arXiv:2505.09388}
}

@article{zeng2025glm,
	title        = {GLM-4.5: Agentic, Reasoning, and Coding (ARC) Foundation Models},
	author       = {Zeng, Aohan and Lv, Xin and Zheng, Qinkai and others},
	year         = 2025,
	journal      = {arXiv preprint arXiv:2508.06471}
}

@misc{anthropic2025claude4,
	title        = {Introducing Claude 4},
	author       = {Anthropic},
	year         = 2025,
	url          = {https://www.anthropic.com/news/claude-4}
}

@article{comanici2025gemini,
	title        = {Gemini 2.5: Pushing the frontier with advanced reasoning, multimodality, long context, and next generation agentic capabilities},
	author       = {Comanici, Gheorghe and Bieber, Eric and Schaekermann, Mike and others},
	year         = 2025,
	journal      = {arXiv preprint arXiv:2507.06261}
}

@article{wei2025browsecomp,
	title        = {Browsecomp: A simple yet challenging benchmark for browsing agents},
	author       = {Wei, Jason and Sun, Zhiqing and Papay, Spencer and others},
	year         = 2025,
	journal      = {arXiv preprint arXiv:2504.12516}
}
